\documentclass[graybox]{svmult}

\usepackage{type1cm}        
\usepackage{hyperref} 
\usepackage{makeidx}         
\usepackage{graphicx}        
\usepackage{multicol}        
\usepackage[bottom]{footmisc}

\usepackage{booktabs,tabularx} 
\usepackage{newtxtext}       %
\usepackage{newtxmath}       

\makeindex             

\usepackage[
backend=biber,
style=numeric,
sorting=anyvt
]{biblatex}
\graphicspath{{figures/}} 
\newcommand{\BB}{B}

\begin{document}

\title*{Machine Learning Assisted Orthonormal Basis Selection for Functional Data Analysis}
\author{Rani Basna, Hiba Nassar and Krzysztof Podg\'{o}rski}

\institute{Rani Basna \at Department of Internal Medicine and Clinical Nutrition, University of Gothenburg, Sweden.  \email{rani.basna@gu.se}
\and
Hiba Nassar \at Cognitive Systems, Department of Applied Mathematics and Computer Science, Technical University of     Denmark, Denmark.  \email{hibna@dtu.dk}
\and Krzysztof Podg\'{o}rski \at Department of Statistics, Lund University, Sweden. \email{krzysztof.podgorski@stat.lu.se}}
%
%
\maketitle

\abstract{In implementations of the functional data methods, the effect of the initial choice of an orthonormal basis has not gained much attention in the past.  Typically, several standard bases such as Fourier, wavelets, splines, etc. are considered to transform observed functional data and a choice is made without any formal criteria indicating which of the bases is preferable for the initial transformation of the data into functions. 
In an attempt to address this issue, we propose a strictly data-driven method of orthogonal basis selection. 
The method uses recently introduced orthogonal spline bases called the splinets obtained by efficient orthogonalization of the $B$-splines. 
The algorithm learns from the data in the machine learning style to efficiently place knots. 
The optimality criterion is based on the average (per functional data point) mean square error and is utilized both in the learning algorithms and in comparison studies. 
The latter indicates efficiency that is particularly evident for the sparse functional data and to a lesser degree in analyses of responses to complex physical systems. 
 }

\vspace{-.9cm}
\section{Introduction}
\label{sec:intro}
\vspace{-.35cm}
This work advances the idea initially presented in \cite{EDOB} and deals with the choice of the initial functional basis selection for functional data analysis (FDA). 
FDA is the field in statistics devoted to the analysis of data that are in the form of functions, images, and shapes, etc. see \cite{ferraty2006nonparametric}. 
The FD may come as a dynamical response from a physical system subject to stochastic excitation that can be written in a generic form as
\begin{equation}
\label{gen}
H(y^{(n)},\dots, y'',y',y,x;\theta)=F(t),
\vspace{-.2cm}
\end{equation}
where $F(t)$ is a realization of stochastic forcing of the system whose response is given by $H$ that involves some physical parameters given in $\theta$.
A classical example here is the harmonic oscillator with damping forced by a Gaussian noise that can be written as 
\begin{align}
\label{eq:damped}
my''+cy'+ky&=dB(t),
\end{align} 
where $m$ represents the mass, $c$ is the viscous damping coefficient, and $k$ is Hook's (spring) constant.
Often the response from such a system is stochastic not only because of random excitation $F$ but also due to randomness in the parameter $\theta$ of the system. As a result, the responses $y_i(t)$ from such a system can be conceptually treated as FD depending both on $\theta_i$ and $F_i(t)$.
The goal is to obtain an efficient treatment of these functional observations in order to infer about $\theta$ as well as about the functional structure of $y$. 

Another type of functional data are sparse data. The sparsity refers to the clear locality of the main features in the data. For example, in the liquid chromatography tandem mass spectra (LC-MS/MS), the data are made of local peaks that are use for identification of metabolomes or proteins.  LC-MS/MS data sets are typically enormous and high dimensional in their nature. If one wants to represent them utilize their sparsity is of the utmost importance. The methods developed in this work target this efficiency which should manifest best in an analysis of such data. 
We present this through simulations of a model that mimics sparsity of the data, while empirical studies involving LC-MS/MS metabolomic data are left for some future research.

In practice, the basis for data representation is often chosen by the mathematical convenience.
Sometimes the number of the basis elements needs to be assessed based on the accuracy of the data representation. Alternatively, large (high-dimensional) bases are used to assure that the representation is sufficiently accurate. In both the cases, one faces inefficiencies leading to unnecessarily high dimensionality of the problem at hand. 
In our approach, we acknowledge the benefit of using spline bases in FDA but we proceed differently from the regularization approach by utilizing knots placement to algorithmic search for efficient knot patterns and use them in the orthogonal basis construction that aims at efficient representation of sparse data. The used construction has been recently proposed in \cite{Liu2019SplinetsE}.  
In the process, we incorporate machine learning algorithms for the choice of basis reducing the mean square error (MSE) uniformly for all samples and study its efficiency against other choices of the basis.  
The optimality criterion is utilized, both in the learning algorithms and in comparison studies.
This criterion allows for comparison performances of different bases in a given problem. 
After efficiently learning from the data about knot placements, we utilize the new construction of the orthonormal spline bases, termed splinets and introduced in \cite{Liu2019SplinetsE} as an efficient orthogonalization of the $B$-splines. 
It has been shown that the splinets bring convenience of the orthogonality while preserving the optimality properties similar to those featured by the $B$-splines.
The method is first tested and illustrated on an artificial example that mimics sparsity in the data since the method would be particularly beneficial to account for sparsity. 
However, we also demonstrate that there is also a gain when we apply the method to a non-sparse and random output from a physical system. 
Namely, we apply the methodology to provide an efficient way of representing the behavior of a stochastic process at the level crossings by providing a simplified functional representation of the so-called Slepian model. 

The paper is organized as follows. We start in Section~\ref{sec:ME} with preliminaries that establish the notation and the main mathematical concepts used in the paper.
There we provide also a brief introduction to the recently proposed orthogonal spline bases, {\it splinets}, which has been implemented in an R-package {\it Splinets}, see \cite{Liu2019SplinetsE} for details on splinets and \cite{ref:Podgorski} for a presentation of the package. 
A motivating example by which we illustrate the main features and benefits of the proposed approach for sparse data in the next section.
Then in Section~\ref{sec:DataDriven}, we present our main contribution which produce machine learning data-driven knot selection for the orthogonal splinet basis.
There we also demonstrate effectiveness of the obtained spline basis by comparing its performance to the one obtained the Fourier basis.
Section~\ref{QVM} is introducing the quarter vehicle model, in which functional responses are modeled through a system of damped harmonic oscillators forced externally by a non-Gaussian form of the noise $d\BB$ occurring at the extreme transient in the road surface. This non-stationary noise is referred to as the Slepian noise and has been discussed in \cite{Podgorski:2015aa}. 
Through this, we illustrate the efficiency of the approach in representation of the model near the extreme event.  The benefits are  evident even if in this case the data are not sparse. 
\section{Preliminaries}
FD are not observed as continuous objects, but high-frequency sampling and mathematical efficiency enable us to see these data as samples of curves, surfaces, or anything else varying over a continuum. 
The fundamental step in FDA is to convert this discrete recorded data to a truly functional form, which allows each function to be evaluated for at any value of its continuous argument. 
In order to utilize the topology of such data for dimension reduction,  one must perform data conversion. 
Typically, one represents a functional object as a linear combination of coefficients and suitable basis functions. 
For this purpose, one of the standard bases such as trigonometric, wavelet, or polynomial is typically chosen. 
Then the efficiency is accomplished by using smoothing through regression or roughness penalty for estimating the coefficients of the basis expansions.
However, all such analyses are preceded by the initial choice of a functional basis used to analyze data, which is hardly objective and is more often driven by mathematical convenience than by the nature of the data itself.  
On the other hand, it is both theoretically and practically observed that the choice of the basis affects efficiency in retrieving the functional structure of a studied model. 

To be more specific, let us consider observations $x_{k}(t)$, $k=1,\dots,n$ that are random elements of $L^2=L^{2}[0,1]$, i.e. the space of square integrable functions on the unit interval. In this Hilbert space, we use inner product $\langle\cdot,\cdot\rangle$ for an integral of the product of its two functional elements and which generates the norm $\| \cdot \|$. 
We call $L^2$-valued data functional observations. We use upper case and lower case letters in the context of FD in a similar manner as in the classical statistical convention, i.e. $X$ is yet not observable random element, while $x=x(\cdot)$ stands for its particular observed functional realization, i.e. a functional outcome of a random experiment carried out according to the probability model for $X$.  
The mean function of the functional data model $X$ is defined by
$$
m(t) = E[X(t)] 
$$
and the covariance function  by
$$
\sigma(s, t) = Cov(X(s), X(t))
$$
for $s,t \in [0,1]$, provided the appropriate expectations exist.
Finally, for the covariance operator on $L^2$ is defined as 
$$
\mathbf K h = \int_0^1 \sigma(\cdot, s) h(s)~ds.
$$

 One can point to a classical result: the Karhunen-Lo$\rm \grave{e}$ve expansion, see \cite{Karhunen}. 
The following theorem shows that the basis associated with this expansion has the optimality in the average mean square error sense, for more details see \cite{hsing}. \vspace{2mm}

\noindent {\it {\bf Theorem (Karhunen-Lo\'eve)} \, For a zero-mean $X(t)$, 
if $Cov(X(t), X(s))=\sigma(t,s)$ is a  continuous function, then there exist a non-increasing square summable sequence of non-negative numbers $\lambda_k$,  an orthonormal basis $e_k,\,{k\in \mathbb N}$  in $L^2[0,1]$ and a sequence of zero-mean variance-one random variables $Z_k$ such that
$$
    {Z}_k = \int_0^1 {X}(t) e_k(t) \, dt/\sqrt{\lambda_k}
$$
and
 \begin{align}
\label{KLM}
{X}(t) = \sum_{k=1}^\infty \sqrt{\lambda_k} Z_k\,e_k(t),
\end{align}
where the convergence is in the mean squared value and is uniform in $t$.
Moreover, the covariance function of the process is represented in the uniform convergence over $[0,1]^2$ as
\begin{align*}
\sigma(s,t)=\sum_{k=1}^\infty \lambda_k e_k(s)\bar e_k(t).
\end{align*}
}

In the spirit of the main data analysis paradigm, for a given FD set it is computationally effective and optimal to work with the basis of the eigenfunctions $\{e_k\}$.  
a data-driven basis and the above set of
However, in the classical functional data, the basis $\{e_k\}$ is the target of the statistical analysis and thus does not serve to represent the data. 
Instead, the data are represented by some other functional basis. 
Consider, for example, the classical smoothing problem, where for a given data we want to fit a smooth function.  Using the $B$-splines together with a regularization method, for example, the Lasso method, one may selectively choose a subspace of the spline space by shrinking parameters to zero, see \cite{GuoHJZ}. Such a basis can be chosen for each FD sample but a choice valid for all samples is not obvious and can significantly affect the accuracy and efficiency of the analysis.

Let us consider $X$ following (\ref{KLM}) with a given choice of $e_k$'s and on the other hand let $(f_k)$ be another orthonormal basis in which the data will be represented, i.e. 
\begin{align}
\label{FDR}
X(t)&=\sum_{i=1}^\infty \langle X,f_i\rangle f_i(t).
\end{align}
We assume that both the orthonormal bases span the same space and for simplicity assume that it is the entire $L^2$.
We have 
$$
\langle X,f_i\rangle = \sum_{k=1}^\infty \sqrt{\lambda_k} Z_k\,\langle e_k, f_i\rangle,
$$
and due to the obvious practical limitations one can only consider the above for a finite number, say $I$, of $f_i$'s.
In this set-up, the criticality of the initial choice of the basis lies both in the fact that we consider only the finite elements of it but also because the basis functions $f_i$, $i=1,\dots, I$, may or may not well approximate $e_k$'s, i.e. the functions
$$
\hat e_k=\sum_{i=1}^I \langle e_k,f_i\rangle f_i
$$
may poorly approximate $e_k$'s. 
Since the eigenvalues $\lambda_i$'s weigh into the quality of approximation it is of importance to choose the basis $\{f_i\}$ so that $\hat e_k$ is close to $e_k$ for large $\lambda_k $. 
 
The proper orthogonal spline bases that are characterized by sparsity understood as local support sets are fundamental for the analysis of sparse functional data. Such bases obtained by appropriate orthogonalization of the $B$-splines have been introduced in \cite{Liu2019SplinetsE}. Here is their short account.

A $B$-spline is a smooth function that consists of polynomial pieces that have the same degree, connected smoothly at join points $\xi_{0}<\xi_{1}<\dots < \xi_{n+1}$, referred to as knots.
The $B$-splines are sensitive to the choice of the knots' position, which is behind our main idea of the basis selection since the choice of the knots can be data-driven. 
Once the knots are set, $B$-splines can be effectively evaluated in a recursive way for any degree using the Cox-de Boor formula \cite{Deboor}. 
The $B$-splines have interesting properties that characterize it. 
Namely, all $B$-splines are positive,  differentiable up to a certain level (the spline order)
and have minimal compact intervals for their supports. 
But except for the case of order zero, the $B$-splines are not orthogonal. 
Different orthogonalization methods appeared in the literature but we are using the structured orthogonalization that creates basis systems referred to as {\it splinet}.

\begin{figure}[t!]
  \centering
  \vspace{0.2cm}
 \begin{minipage}{0.495\textwidth}
\includegraphics[width=1\textwidth,height=0.65\textwidth]{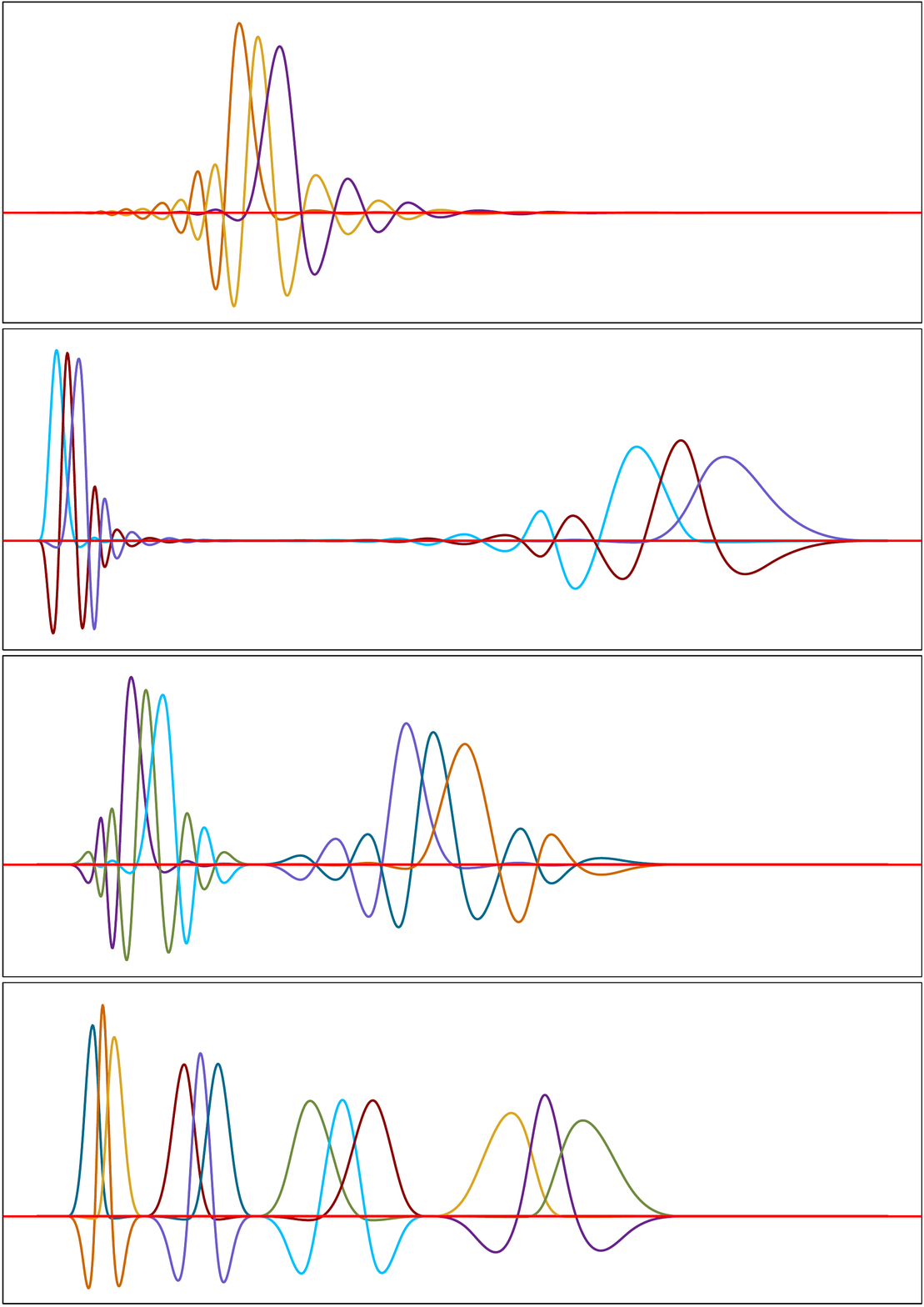}
\end{minipage}
\begin{minipage}{0.495\textwidth}
\includegraphics[width=1\textwidth,height=0.65\textwidth]{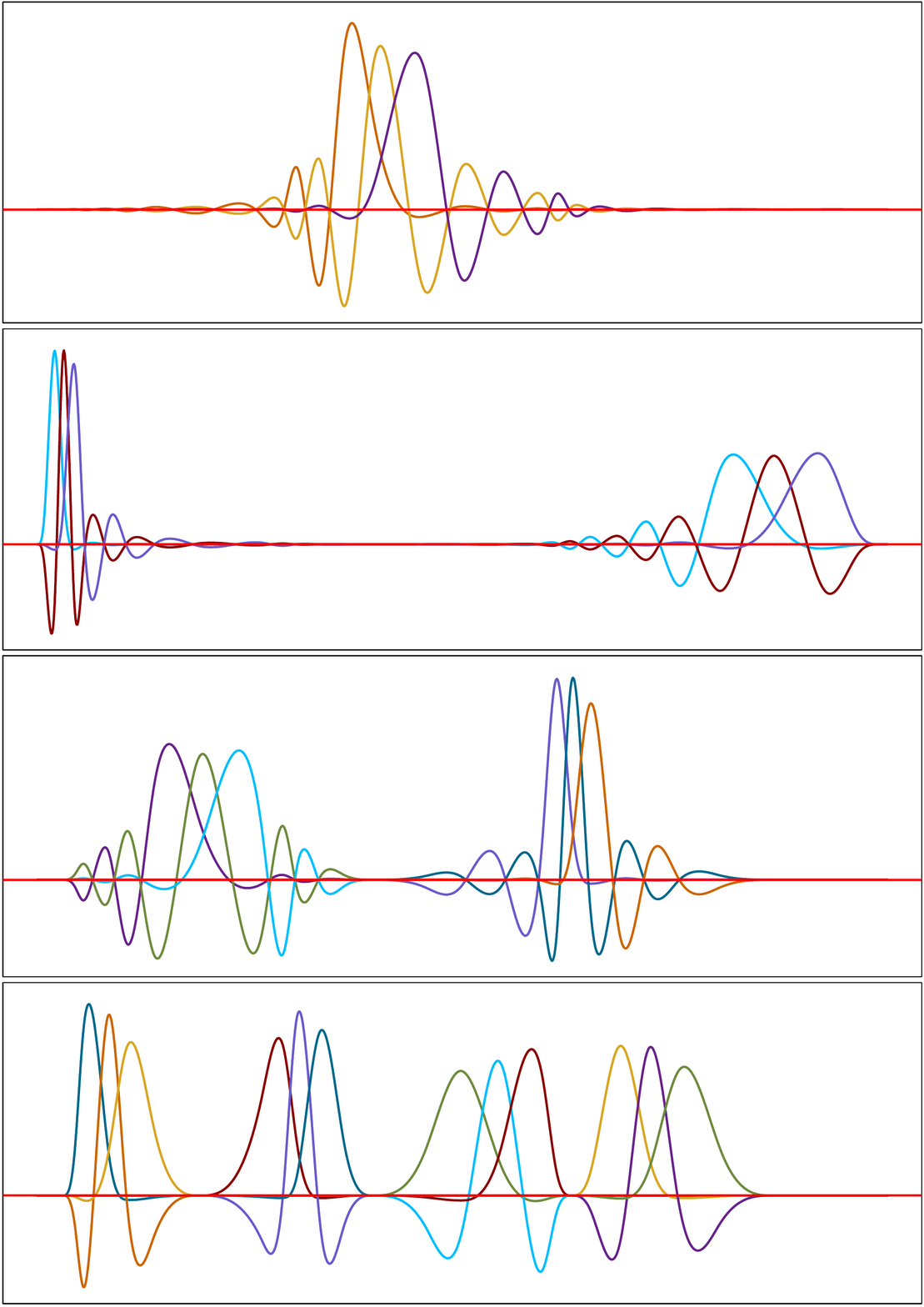}
\end{minipage}
  \caption{Two Splinets for the data driven knots placement for examples of Subsection~\ref{subsec:eff}. The locality and sensitivity to the knot placement is evident.   }
  \label{fig:OB}
  \vspace{-0.5cm}
\end{figure}

The splinet is prioritized over other orthonormal spline systems as it preserves locality and computational efficiencies of the original splines.  For a more detailed explanation, we refer the reader to \cite{Liu2019SplinetsE}.
Given the knots, these spline bases are obtained from the $B$-splines through efficient dyadic orthogonalization that is performed in a self-similar fashion and thus preserving the locality entertained by the $B$-splines. 
In Figure~\ref{fig:OB},  we see the ability to represent the local detail of two splinets spread over two different knot placements. 

\section{A motivating sparse data example}
\label{sec:ME}
To illustrate the basic features of the proposed approach throughout the paper, we consider (\ref{KLM}) that follows the  general finite dimensional set-up.
For a certain generic orthonormal basis $(f_i)$, $i=1,\dots, I$, the eigenfunctions $(e_k)$, $k=1,\dots, K$   are defined through
\begin{align*}
e_k= \sum_{i=1}^I a_{ki}f_i
\end{align*} 
where the $K\times I$ matrix $\mathbf A=(a_{ij})$ satisfies 
$$
\mathbf A \mathbf A^\bot = \mathbf I_K,
$$
where $\mathbf I_K$ is the $K\times K$ identity matrix. 
Further $\lambda_k$, $k=1,\dots,K$ are non-increasing  eigenvalues  corresponding to $e_k$.

\begin{figure}[t!]
\begin{center}
\includegraphics[width=0.9\textwidth,height=0.4\textwidth]{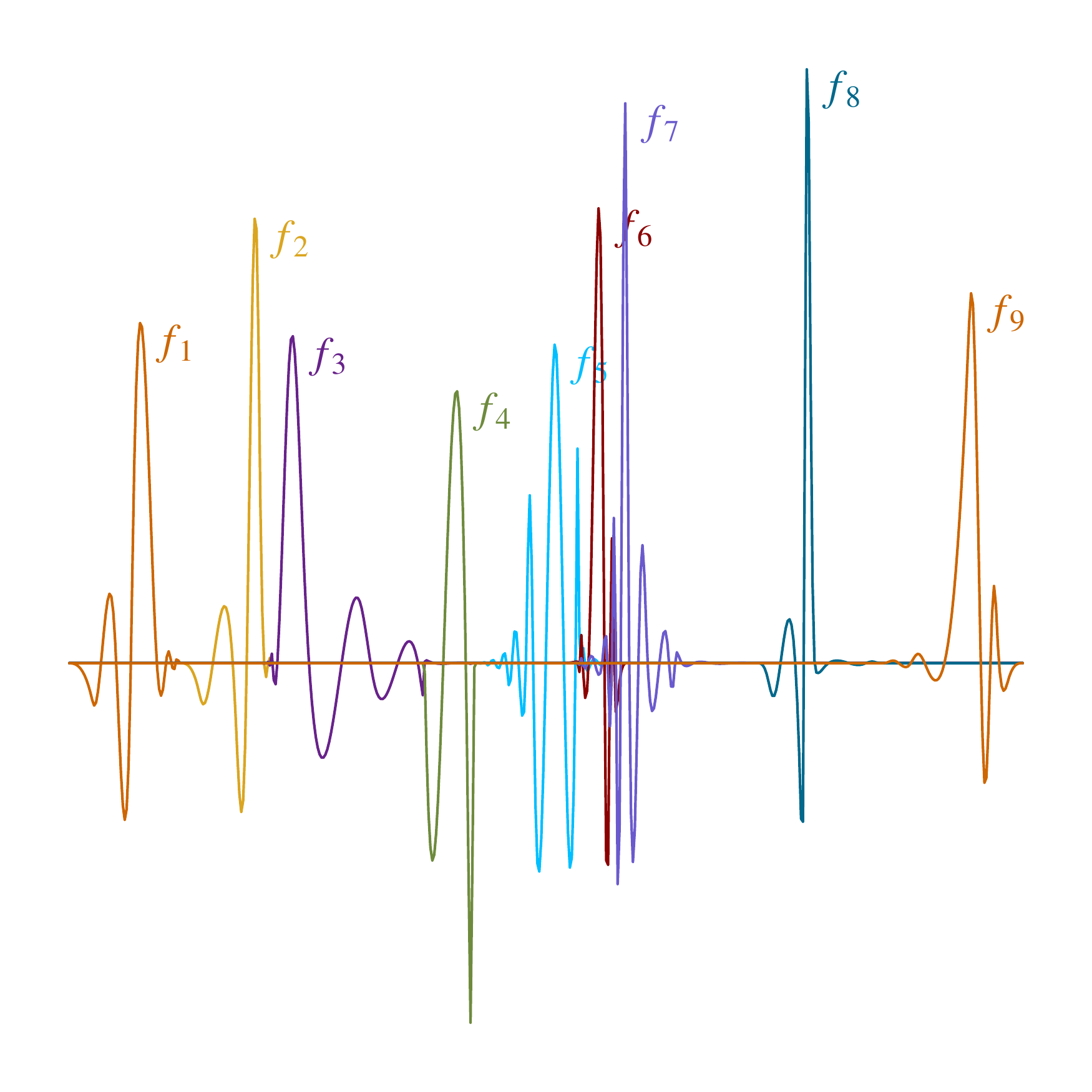}\\
\includegraphics[width=0.9\textwidth,height=0.4\textwidth]{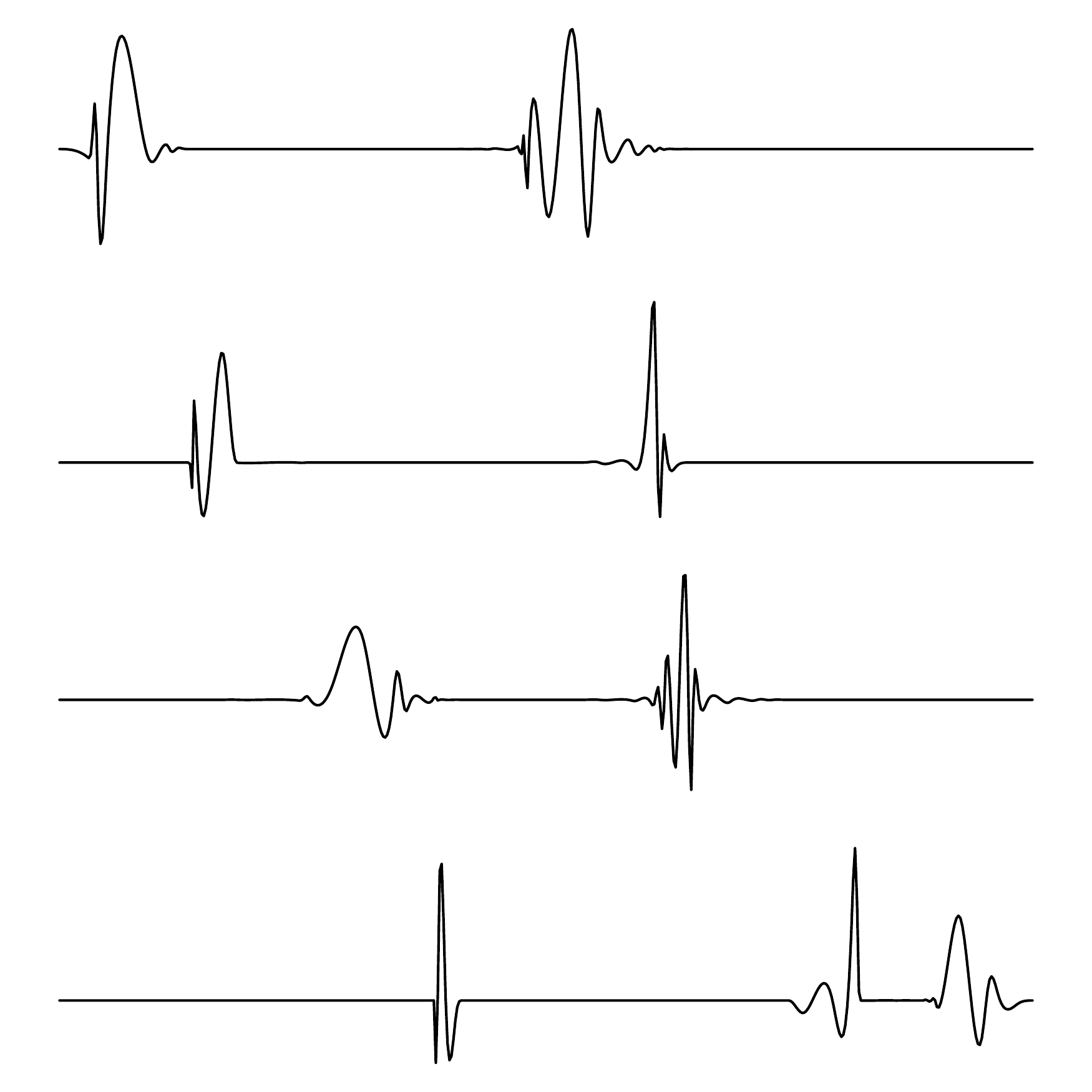}\vspace{-1cm}\\
{\includegraphics[width=0.9\textwidth,height=0.7\textwidth]{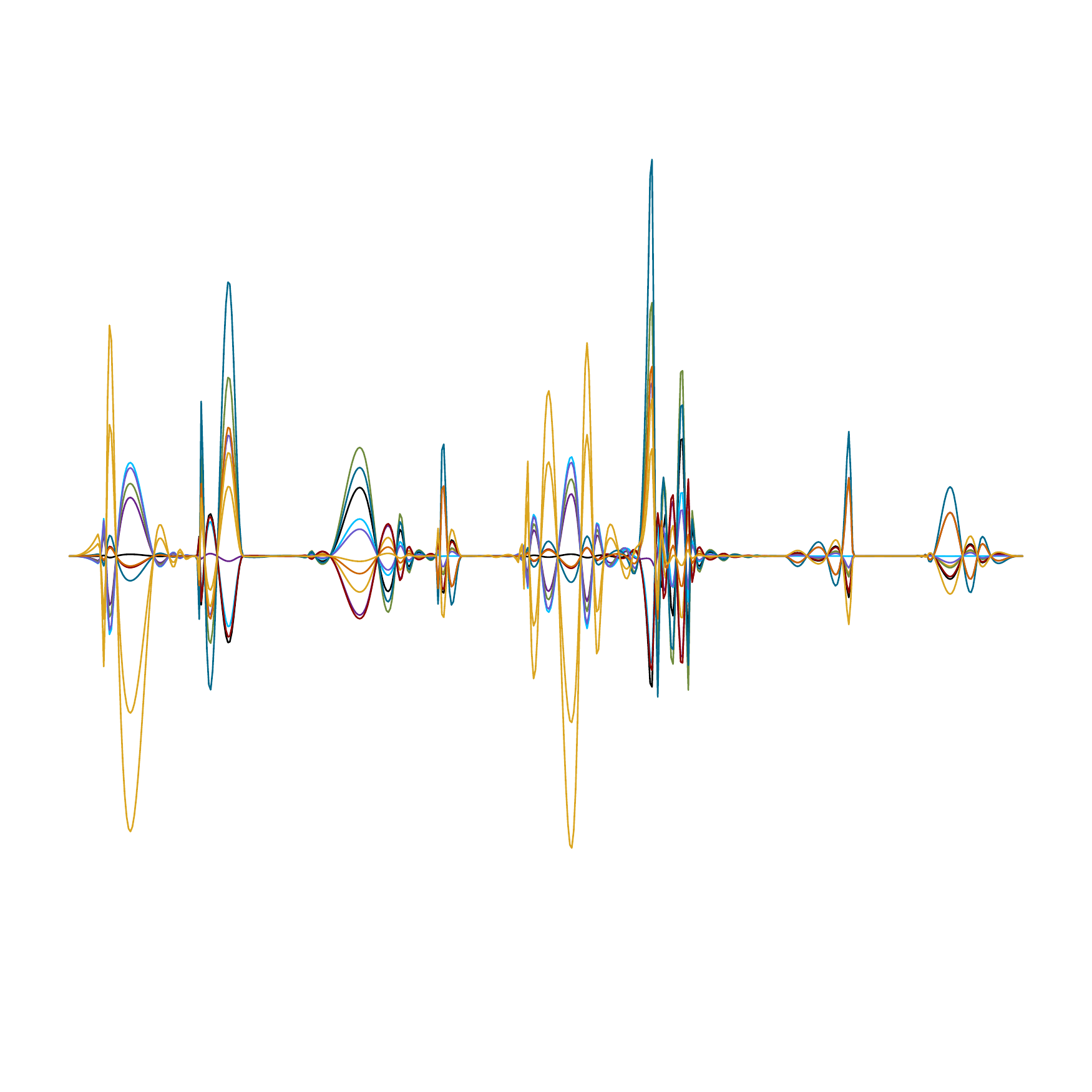}} \vspace{-2cm}
\end{center}
\caption{The basis $\{f_i\}$, $i=1,\dot,9$ giving a complete representation of the data  (top). The eigenfunctions $\{e_k\}$, $k=1,\dots, 4$ (middle) for the functional model (\ref{KLM}). 
Ten samples from the model are shown in the bottom graph. }
\label{IllEx}
\end{figure}

 In a simple specification of the above model, we take $I=9$ and for $(f_i)$, $i=1,\dots, 9$ we take the third order orthogonal splines that are elements of a splinet spanned on irregularly placed knots, see \cite{Liu2019SplinetsE}. 
Due to the sparsity of the $B$-splines that is inherited by the splinet, the model serves as a toy example of sparse data generator. 
The splinet (orthonormal functional basis) is presented in  Fig.~\ref{IllEx}, the top graph. 
We take $K=4$ eigenfunctions defined through the matrix $\mathbf A$ given in 
$$
\mathbf A = 
\begin{bmatrix}
2^{-\frac 12}& 0 & 0 & 0 & 2^{-\frac 12} & 0& 0 & 0 & 0\\
0& 2^{-\frac 12} & 0 & 0 & 0 & 2^{-\frac 12}& 0 & 0 & 0\\
0& 0 & 2^{-\frac 12} & 0 & 0 & 0& 2^{-\frac 12} & 0 & 0\\
0 & 0 & 0 & 3^{-\frac 12}& 0 &0& 0 & 3^{-\frac 12} &3^{-\frac 12}
\end{bmatrix}
$$
that leads to the normalized eigenfunctions $e_k$, $k=1,\dots,4$,  shown in Fig.~\ref{IllEx} the middle graph. 
The four corresponding eigenvalues are 
$$
(\lambda_1,\lambda_2,\lambda_3, \lambda_4)=(1, 0.5, 0.3, 0.01).
$$
Ten samples from \eqref{KLM} in which we assumed $Z_i$'s to be the standard normal variables, the case of a Gaussian model, are shown in the bottom graph of Fig.~\ref{IllEx}.

In this simple example, one can illustrate how critical is the initial choice of a basis used for the representation of the data. 
For example, if one chooses the basis $\mathcal B_1=\{f_3,\dots,f_9\}$, then the mean square error of using this basis for the data representation is
\begin{align*}
E\left\|\sum_{k=1}^K \sqrt{\lambda_k} Z_k\,\left(e_k(t)-\hat e_k(t)\right)\right\|^2&=E\left\|\sum_{k=1}^K \sqrt{\lambda_k} Z_k\,\left(a_{k1}f_1+a_{k2}f_2\right)\right\|^2\\
&=E
\left\|
\sum_{k=1}^K \sqrt{\lambda_k} Z_k a_{k1}f_1
\right\|^2
+
E\left\|
\sum_{k=1}^K \sqrt{\lambda_k} Z_ka_{k2}f_2
\right\|^2
\\
&=E
\left(
\sum_{k=1}^K \sqrt{\lambda_k} Z_k a_{k1}
\right)^2
+
E\left(
\sum_{k=1}^K \sqrt{\lambda_k} Z_ka_{k2}
\right)^2
\\
&=
\sum_{k=1}^K {\lambda_k}\left( a_{k1}^2+ a_{k2}^2
\right)
\\
&=0.85.
\end{align*} 
which leads to major inaccuracies.
On the other hand similar computations for  the basis $\mathcal B_2=\{f_1,\dots,f_7\}$ lead to the error $0.046$ which is rather negligible as compared with the total mean squares norm of the data which is $\lambda_1^2+\dots + \lambda_4^2=1.3401$ . 
Of course, choosing the full basis $f_1,\dots,f_9$ will yield the exact representation of the data. However, in practice, such an exact representation rather should not be expected. Firstly, it is typically assumed that the data are from infinite dimensional space so that there are infinite many $e_i$'s in \eqref{KLM}, while an analyst will have only a finite number of the basis elements available. 
The further complication of the problem is that the mean square error can be only estimated and further confounded by the noise in the data. 

We note that if $\sigma(t,s)$ is available, then in this finite dimensional case the eigenvalues $\lambda_i$ and the eigenfunctions $e_i$ are simply found by finding the eigenvalues and eigenvectors of the matrix. 
\begin{equation}
\label{Sigma}
\mathbf \Sigma = \left[ \langle f_i, \mathbf K f_j\rangle \right]_{i,j=1}^I.
\end{equation}
The non-zero eigenvalues correspond to $\lambda_i$'s while the corresponding eigenvectors $\mathbf e_i\in \mathbb R^9$ if normalized represent $e_i$ through
$$
e_k(t)=\sum_{i=1}^Ie_{ki}f_i(t).
$$
However, in the case when $\sigma$ and thus $\mathbf K$ is only estimated from the data by $\hat \sigma$ and $\hat K$, respectively, the issue of the choice of the initial basis, in addition to the sample size, becomes a sensitive factor for the accuracy of the eigenvalue/eigenfunction estimation.

%
In a more realistic setup, the observed data are also involving an observational noise with some given standard deviation $\sigma_0$. Thus in a simple model that accounts for the noise, we consider
 \begin{align}
\label{KLMwithnoise}
{X}(t) = \sum_{k=1}^K \sqrt{\lambda_k} Z_k\,e_k(t) + \sigma_0 dB(t),
\end{align}
where $B$ is a Brownian motion independent of $Z_k$'s. 
Consequently, for any $f_i$, $i=1,\dots,I$, we have the following projection of the data into the basis $(f_i)$:
$$
\langle \mathbf X, f_i\rangle = \sum_{k=1}^K \sqrt{\lambda_k}a_{ki} Z_k +\sigma_0 \int_0^1 f_i dB(t).  
$$
We observe that the noise compoenent $\epsilon_i=\sigma_0 \int_0^1 f_i dB(t)$ has the variance $\sigma^2_0$ one and constitutes of uncorrelated random variables. 
This means that instead of the covariance  given in (\ref{Sigma}) the coefficients of the data expansions in the basis $(f_i)$ have the covariance 
\begin{equation}
\label{cov}
\boldsymbol \Sigma_1=\boldsymbol \Sigma + \sigma_0^2 \mathbf I_{I},
\end{equation}
where $\mathbf I_{I}$ is the $I\times I$ identity matrix.
In this situation, an efficient basis is beneficent not only because of its computational convenience but it allows also to reduce the estimation error if the sample size is small comparing to the dimension of the basis. 

\begin{figure}[t!]
\begin{center}
\includegraphics[width=0.8\textwidth]{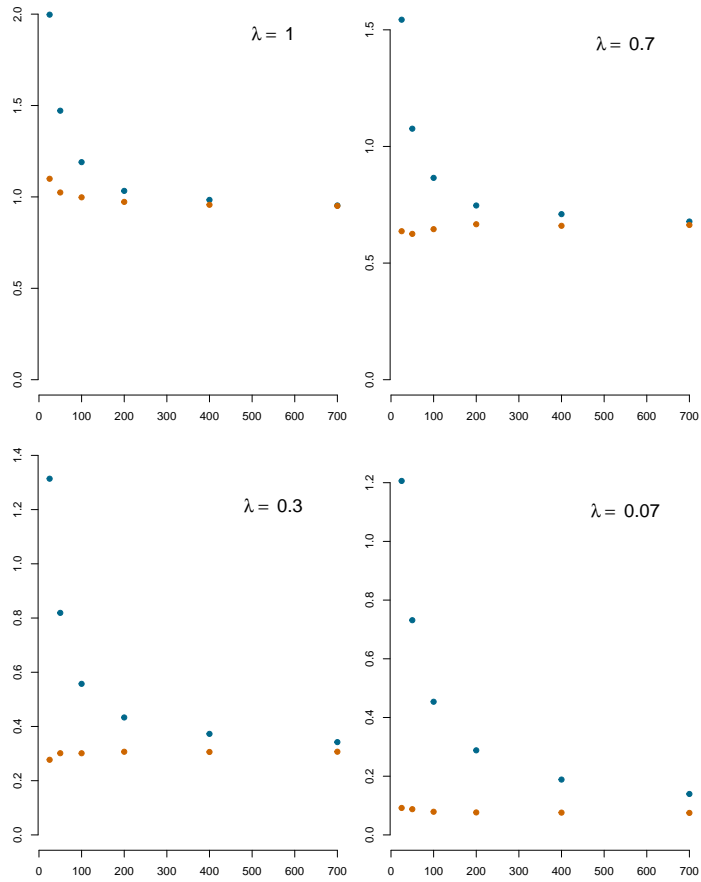}\\
\end{center}
\caption{ 
  Mean of the four eigenvalues for the estimated covariance matrix obtained from $200$ Monte Carlo simulations.  Two different cases of the initial basis selection, the case $f_i$, $i=1,\dots,9$ (orange-lower dots), and 200 splinets $OB_i$,$i=1,\dots,200$ (blue-upper dots) as a function of the base size, which ranges from 25 to 700.}
\label{mean1vsmean2}
\end{figure}

Next, we study  the effect of the sample size on precision of the eigenvalue/eigenfunction estimation based $\hat \sigma$ in two cases of the initial selection of the orthogonal bases:  $f_i$, $i=1,\dots,9$, and 200 splinets $OB_i$, $i=1,\dots,200$ created on equally spaced knots.  We consider the case of noisy data with the variance $\sigma^2_0 =0.1$ and 6 different sample sizes: $25,50,100,200,400,700$.  For each sample size, we run a Monte Carlo study with $MC=200$ to compute the eigenvalues of the covariance matrix \eqref{cov} for the two initial choices of a basis. Figure~\ref{mean1vsmean2} shows that the four eigenvalues of the estimated covariance matrix make good estimators for the actual eigenvalues in the case of  $f_i$, $ i=1,\dots,9$ as the initial basis despite the sample size, in contrast to the case of 200 splinets $OB_i$, $i=1,\dots,200$ where eigenvalues of the covariance matrix become good estimators for the actual eigenvalues only when the sample size is fairly big.

Moreover, the Monte Carlo study also shows that the mean square errors (MSE) of the estimated eigenvalues for small sample sizes are significantly better if the `correct' basis of e$f_i$, $i=1,\dots,9$ is chosen instead of 200 splines with equally spaced knots, see the Table~\ref{MSE}.

\begin{center}
\begin{table}
\resizebox{\columnwidth}{!}{
 \begin{tabular}{c||c c c c || c c c c} 
 \hline
 {\small Sample Size} & {\small  MSE$(\lambda_1)$ }&  {\small MSE $ (\lambda_2)$ }&  {\small MSE$ (\lambda_3)$ }& {\small MSE$ (\lambda_4)$ }& {\small MSE$ (\lambda_1)$ }& {\small  MSE$(\lambda_2)$ }&  {\small MSE$ (\lambda_3)$ } & {\small MSE $ (\lambda_4)$ }  \\ [0.5ex] 
 \hline\hline
 25&0.78& 0.20 & 0.10 & 0.04 & 2.58& 0.94& 1.10& 1.32 \\ 
 \hline
 50&0.57 & 0.14& 0.07 & 0.03 &1.14 &0.31& 0.29& 0.46 \\
 \hline
100& 0.43 &0.11& 0.06& 0.02 & 0.69 &0.14& 0.09& 0.16 \\
 \hline
 200&0.37& 0.08& 0.04 & 0.02 & 0.48& 0.08 &0.04& 0.06 \\
 \hline
 400&0.31& 0.05& 0.03 &0.01 & 0.35& 0.05& 0.02& 0.02\\  
 \hline
 700& 0.29 & 0.04& 0.02& 0.01 & 0.31 &0.04 &0.02 &0.01 \\ [1ex] 
 \hline
\end{tabular}
}
\caption{ MSE of the estimated eigenvalues in case of $f_i$, $i=1,\dots,9$ (left) and in case of $200$ splinets (right). }
\label{MSE}
\end{table}
\end{center}
Our illustrative example emphasizes the importance of choosing the initial basis. Of course, in reality, it is not possible to know a priori the basis that generates the model. Now, we turn to a method of selecting a data-driven orthogonal basis.

\section{Data driven choice of the knots}
\label{sec:DataDriven}
\vspace{-.35cm}
The common degree of polynomials and the placement of knots define the spline basis. 
While the order 3 is the most popular generic choice, often the degree of the involved polynomials may be decided by the nature of the data.  
For example, if one considers the Brownian bridge $B$, then the samples are continuous but not differentiable and the first order splines (piecewise linear functions) is a natural choice. 
On the other hand, if one chooses the Laplace bridge $L=B\circ \Gamma$, where $\Gamma$ is a gamma process, which is an example of a pure jump process, using the zero order splines may be the most natural to represent process samples. 
Overall, the choice of the spline order is a separate issue and we do not discuss it in any further detail here. 
Our focus is on the choice of knots, which is at the center of the proposed methodology. 

\begin{figure}
  \includegraphics[width=\linewidth]{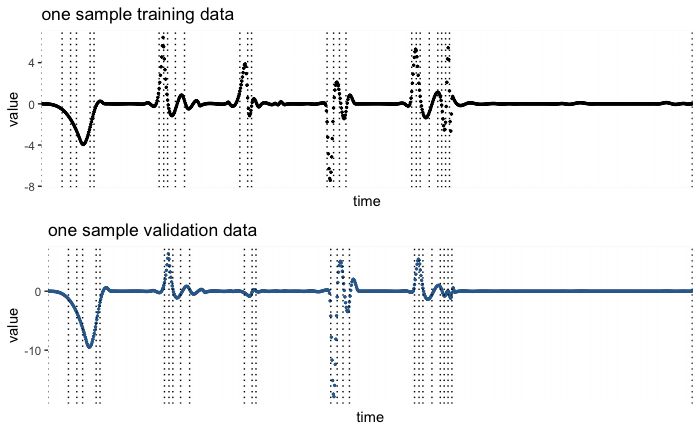}
  \caption{({\it top}) One sample from the train set of the motivated example model. ({\it bottom}). One sample from the validation set of the motivated example model. The vertical dashed lines refer to the 25 data-driven knots selected with validation. The knots emerged from using the first criterium with predefined step size $\mathcal\theta = 0.38$.}
  \label{fig:splits}
\end{figure}

In this work, we propose machine learning style techniques for the placement of the knots. 
The chosen knots are used to build orthogonal splines basis functions $f_k (t)$, $k=1,\dots, K$ that are used in basis function expansion to convert the data from discrete recorded data into a functional one.
We use a standard random split of the data to {\it train} and {\it validation} parts, although other machine learning techniques such cross-validation, bagging, etc. can be utilized for the purpose as well. The random splitting preserves the original discretization for each sample. More specifically, a functional data set $\mathcal X$, is divided randomly to $ \mathcal X_{train}$ and $\mathcal X_{valid}$, such that, each sample $x_i$ belongs to one of the sets $ \mathcal X_{train}$ or $\mathcal X_{valid}$. For the current presentation, we choose the size of the data splits to be $60\%$ as training, $40\%$  as validation. Figure~\ref{fig:splits} is an example of one sample from the sparse data example model, Section~\ref{sec:ME}, of  2000 observations, divided with the previous percentage. We will call the method Data Driven Knots (DDK) and its initial implementation in R language is available as a package at the \href{https://github.com/ranibasna/ddk}{GitHub page: {\tt https://github.com/ranibasna/ddk}}. The package will be under continuous development.


\subsection{Training}

The method of adding knots is based on the mean square error effectiveness of approximating the FD. 
The method is iterative and resembles the regression tree building, see \cite [Chapter 9]{HastieTF9}. 

For any FD set $\mathcal X=\{x_i\in L^2, i=1,\dots n\}$, the set of best least square constant predictors is a set of functions \vspace{-.2cm}
$$
x_i^{(0)} =\langle x_i, \mathbf 1\rangle \mathbf 1= \int x_i\cdot \mathbf 1.
\vspace{-.2cm}
$$ 
The constant functions over the entire domain $[0,1]$ can be viewed as $0$-order splines with no internal knot points, and its one dimensional basis is given by the constant function $\mathbf 1$.
We set the initial set of knots to an empty set, i.e. $\mathcal K^{(0)}=\emptyset$, the initial basis $\mathcal B^{(0)}=\left\{\mathbf 1\right\}$, and the projection to the space spanned by $\mathcal B^{(0)}$ is given by $\mathbf P^{(0)} x=\langle x, \mathbf 1\rangle \mathbf 1$.  
The average mean square error (AMSE) per function of the approximations of $x_i$'s by the optimal constant functions is given by\vspace{-0.2cm}
$$
AMSE(\mathcal Y,\mathcal B^{(0)})=\frac{1}{n}\sum_{i=1}^n \|\ x_i - \mathbf P^{(0)} x_i\|^2=\frac{1}{n}\sum_{i=1}^n \|\ x_i - \langle x_i, \mathbf 1\rangle \mathbf 1\|^2.\vspace{-.3cm}
$$

The method at the first step, $s=1$, finds a knot $\xi\in [0,1]$ such that the optimal approximation of $x$ by a linear combination of the $0$-order splines with the set of knots $\mathcal K^{(1)}=\mathcal K^{(0)}\cup \{\xi \}$ yields the smallest AMSE between the FD $x_i$.
In other words, denote by $\mathcal B^{(1)}(\xi)$ the orthonormal basis of piecewise constant functions over the intervals given by the knots in $\mathcal K^{(1)}(\xi)$.
The new knot $\xi_{new}$ is chosen as
\begin{equation}
\label{new}
\xi_{new}= \underset{\xi \in (0,1]}{\rm argmin~}AMSE(\mathcal Y,\mathcal B^{(s)}(\xi) ).
\vspace{-.3cm}
\end{equation}
Then the new, enlarged by one function, basis $\mathcal B^{(1)}=\mathcal B^{(1)}(\xi_{new})$ is uniquely defined by the new set of knots $\mathcal K^{(1)}=\mathcal K^{(1)}(\xi_{new})$. In the recurrent process, at the step $s$, we start with a sequence of knots $\mathcal K^{(s-1)}$ and search for a new knot $\xi_{new}$ using (\ref{new}) with $\mathcal K^{(s)}(\xi)=\mathcal K^{(s-1)}\cup \{\xi \}$
and the corresponding orthonormal basis of piecewise constant functions $\mathcal B^{(s)}(\xi)$. 

The algorithm benefits from the locality and orthogonality piecewise constant bases so that each new knot requires a removal of only one base function (the constant over an interval that includes the new knot) and replaces it with two new functions that remain orthonormal to all the other bases functions from the previous step. 

\subsection{Validation}
The number of selected knots plays an essential role. A large number of knots may result in overfitting. In contrast, a small number may result in underfitting the
data. To address this problem, a validation index using a stopping threshold has been proposed motivated by traditional machine learning algorithms. Namely, we developed stopping criteria that automate the convergence time of the above-explained knots selection method.  
For the stopping threshold, we suggest two criteria: the step difference between consecutive
AMSE values and the relative step difference between consecutive AMSE values. Both criteria are achieved by using a predefined step size threshold. The step size at iteration $s$ is defined as the absolute difference between two sequential average mean square error values,   $ | AMSE_{s} - AMSE_{s+1}| $, where AMSE$_{s}$ is the value of the AMSE at the step $s$. The step size threshold $\theta$ is a bound on the value of a step the AMSE function takes.

The validation procedure is running simultaneously to the training one. At the iteration step $s$, the training algorithm delivers the new set of knots $\mathcal K^{(s)}$ and a new set of pieces-wise constant basis $\mathcal B^{(s)}(\xi)$.  In the next step,  the AMSE$(\mathcal Y_{valid},\mathcal B^{(s)}(\xi)) $  is computed and two criteria for the stopping: the step difference between consecutive AMSE values and the relative step difference between consecutive AMSE values.

By the first criterium, if the algorithm at iteration $i$ reduces the AMSE value by a value smaller than the threshold $\theta$ the iteration stopped. No more knot selection is performed. In other words, we stop the iteration if  the following condition is met
$$  \mid AMSE_{s} - AMSE_{s-1} \mid  < \theta.$$ 

Similarily, for the second criterium, the algorithm stops at iteration $s$ if the relative step difference between consecutive AMSE values is smaller than the threshold $\theta$. The relative step difference is the ratio of  the absolute difference to a reference value
$$  \mid AMSE_{s} - AMSE_{s-1} \mid < \mathcal\theta  \mid{AMSE_s}\mid.$$ 

It is clear that in mathematical terms, the increase of the number of knots is no longer needed when the knots start to fit a pure noise instead of the actual functional structure.   
In the future development, we plan to expand these ideas to a full-blown statistical test that will analyze the behavior of the above criteria when they are applied to data sampled from $dB$, where $B$ is a Brownian motion: the algorithm should stop adding knots when the hypothesis that one deals with a pure noise cannot be rejected.

For our motivated example of Section~\ref{sec:ME}, we run the validation process using the first criterium with predefined steep size $\mathcal\theta = 0.38$. The process resulted in 25 iterations as the optimal stopping number of iterations. Figure \ref{fig:splits} displays the outcome of the validation process and the selected knots from the training and validations represented in dashed lines.

\begin{figure}
  \includegraphics[width=\linewidth]{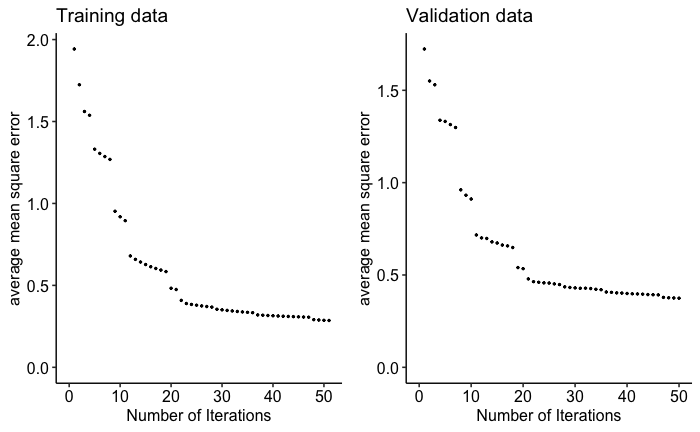}
  \caption{Reduction in AMSE in the motivated example of section \ref{sec:ME}. {\it Left}: reduction in AMSE achieved after each additional knot selection during training. {\it Right}: reduction achieved after each additional knot selection on the validation data.}
  \label{fig:train_validate}
\end{figure}

 Figure \ref{fig:train_validate} shows both the AMSE evaluation on $\mathcal X_{train}$ and $\mathcal X_{valid}$. It is evident that in the training phase each knot selection imposes a reduction on AMSE with different step sizes. On the validation data, most of the time the newly added knot helps in reducing the AMSE.  Few times, the newly selected knot does not contribute to the AMSE reduction. This is partially due to the random splitting where some intervals have more training data than validation ones. Overall, Figure~\ref{fig:train_validate} explains the mechanism of stopping the iteration process. For the motivated example model of section \ref{sec:ME}, depending on the predefined value of $\theta$, the iteration may stop in the vicinity of the 25th iteration.
\subsection{Efficiency in the FD representation}
\label{subsec:eff}
To illustrate gains in the FD representation using the DDK-splines we consider a very simple example of functional data and decompose them first using the Fourier basis and then the DDK-splines.

The Fourier basis is probably the most popular basis used for a decomposition of a function. 
It consists of the following functions
$$
\{{\sqrt {2}}\sin(2\pi nt); \;n\in \mathbb {N} \}\cup \{{\sqrt {2}}\cos(2\pi nt); \;n\in \mathbb {N} \}\cup \{\mathbf 1\}
\vspace{-.2cm}
$$
and form an orthogonal basis.  A Fourier decomposition is especially useful for smooth functions where there are no strong local features present and the comparable curvature order everywhere. However, they are improper for data where discontinuities in the function itself or in low order derivatives are known or suspected \cite[page 48]{ramsay2004functional}.

\begin{figure}[t]
\begin{center}
\vspace{-0.3cm}
\includegraphics[width=0.24\textwidth]{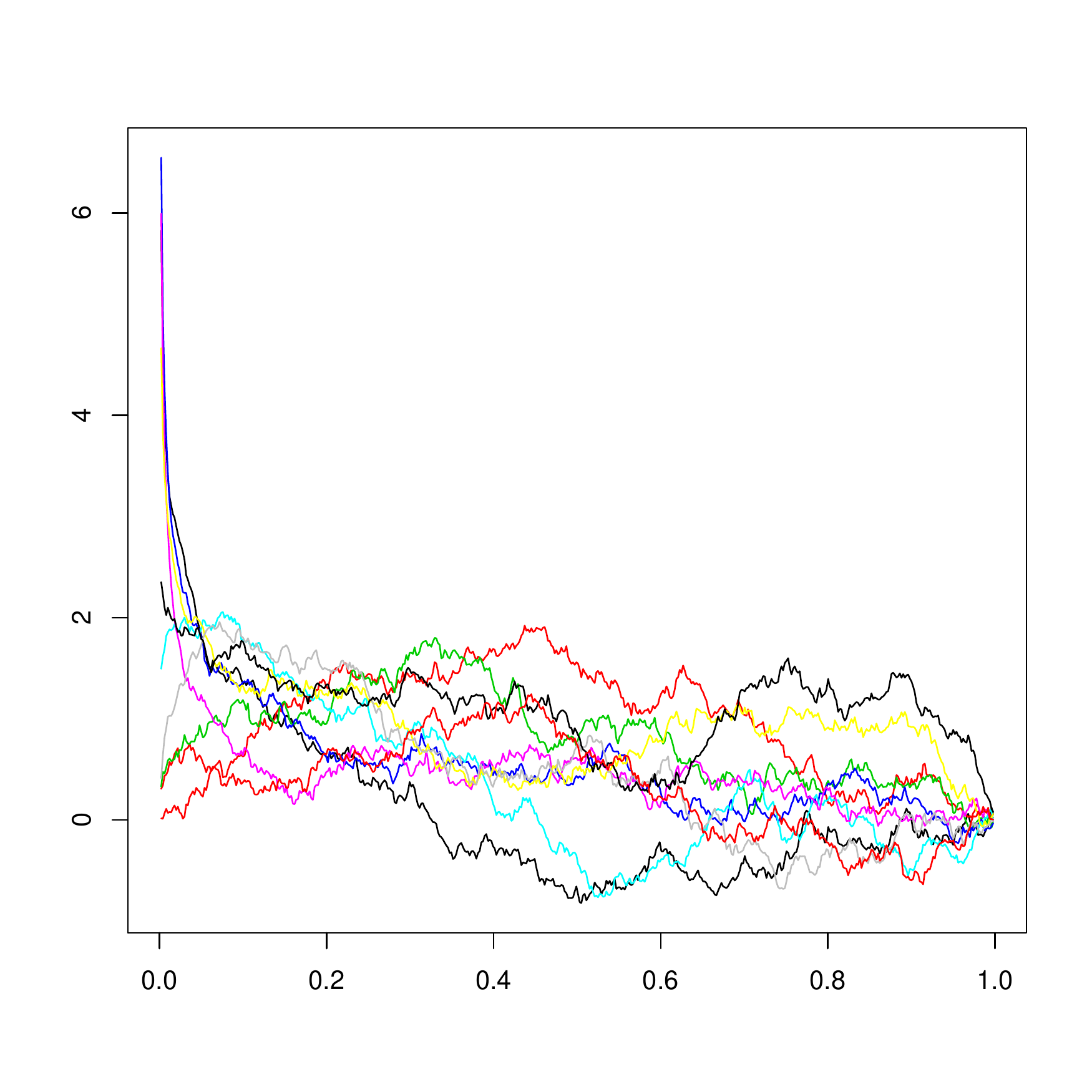} 
\includegraphics[width=0.24\textwidth]{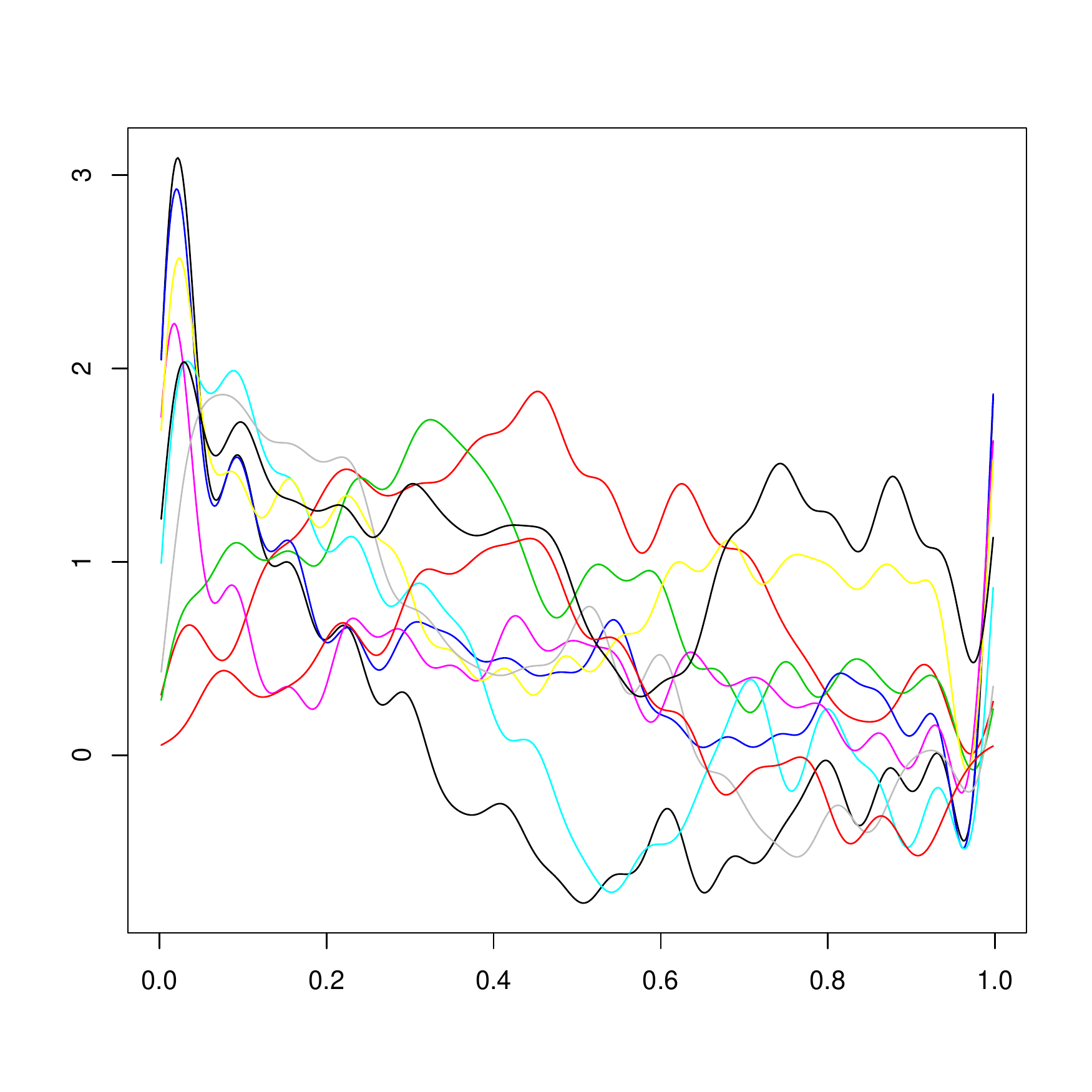}
\includegraphics[width=0.24\textwidth]{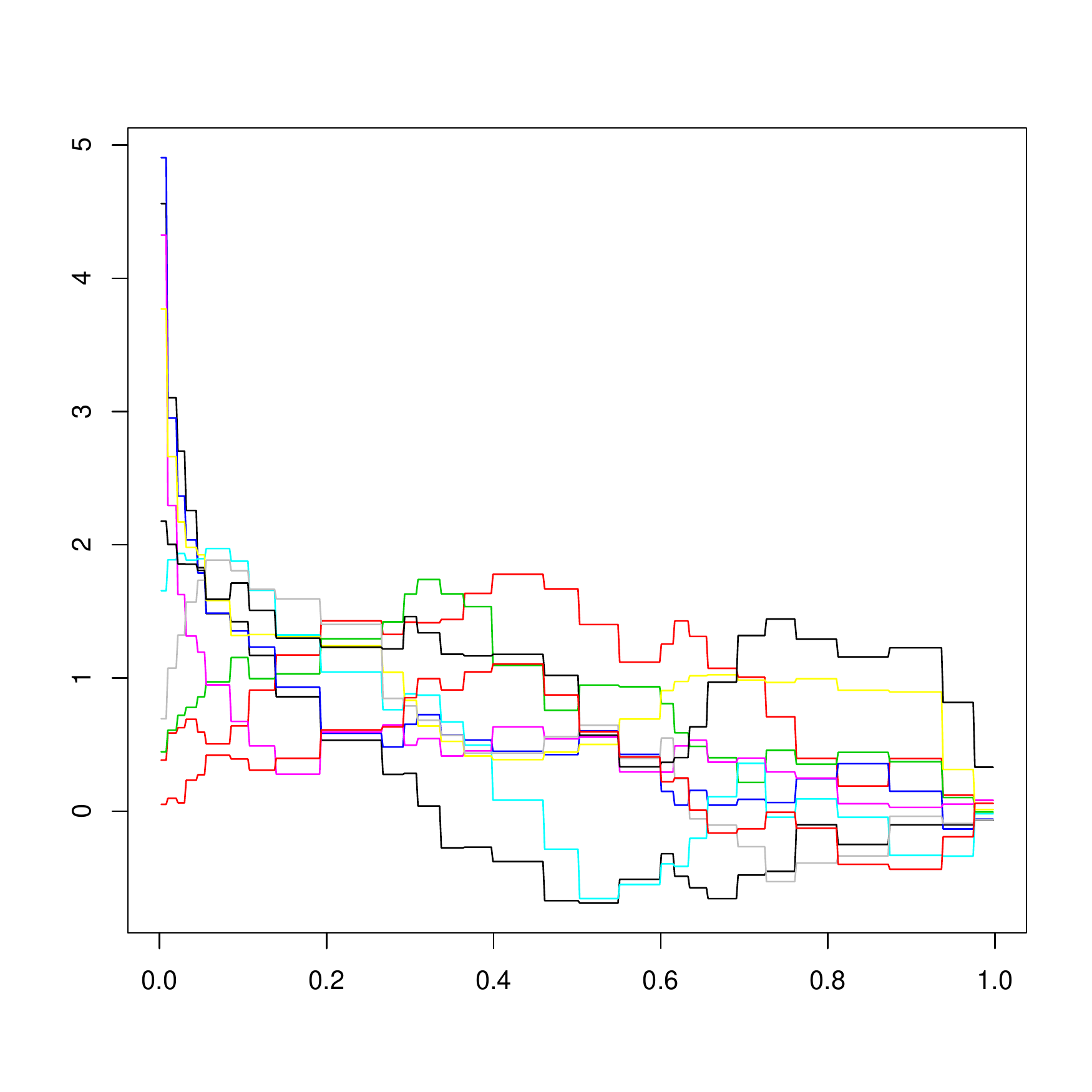} 
\includegraphics[width=0.24\textwidth]{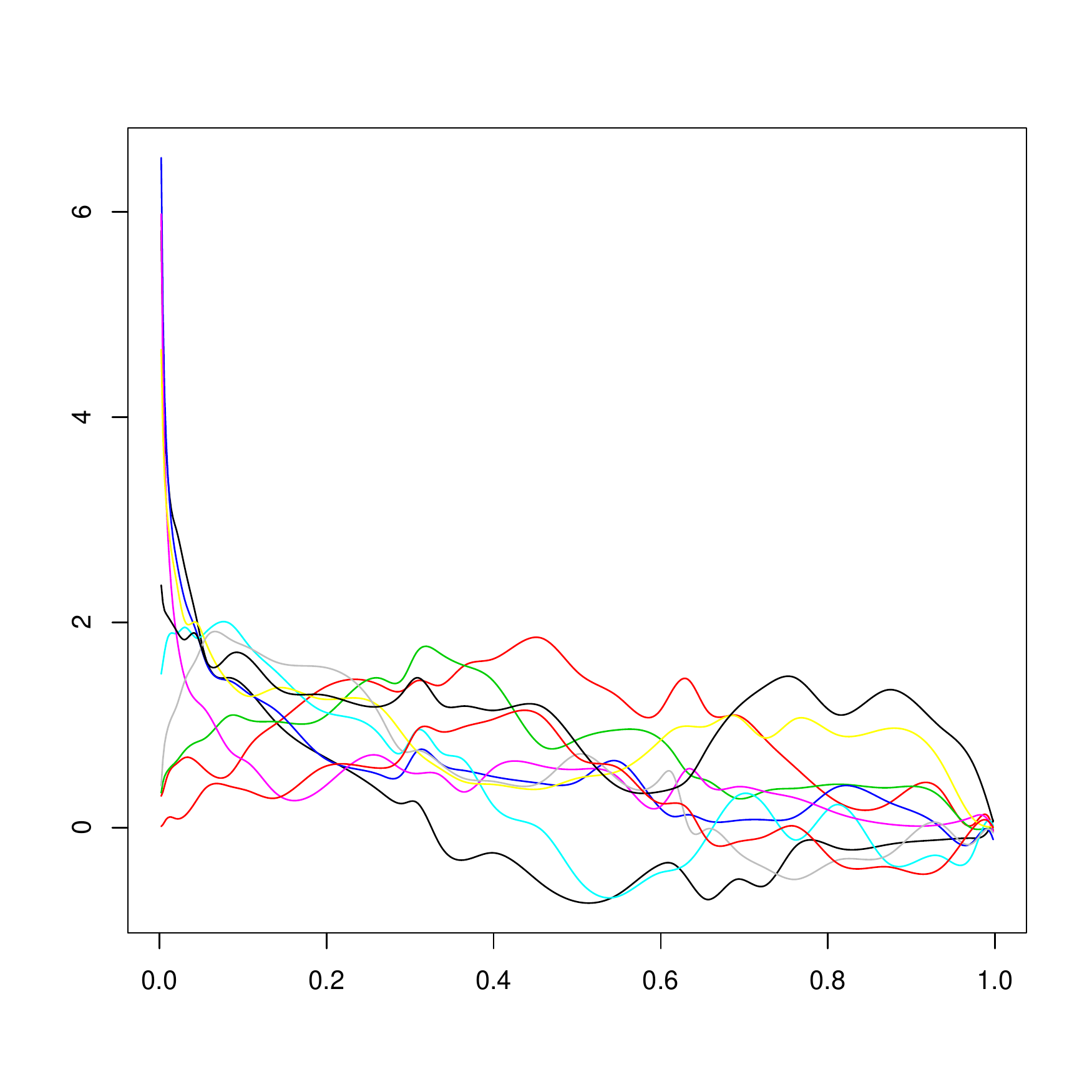} \\
\includegraphics[width=0.49\textwidth]{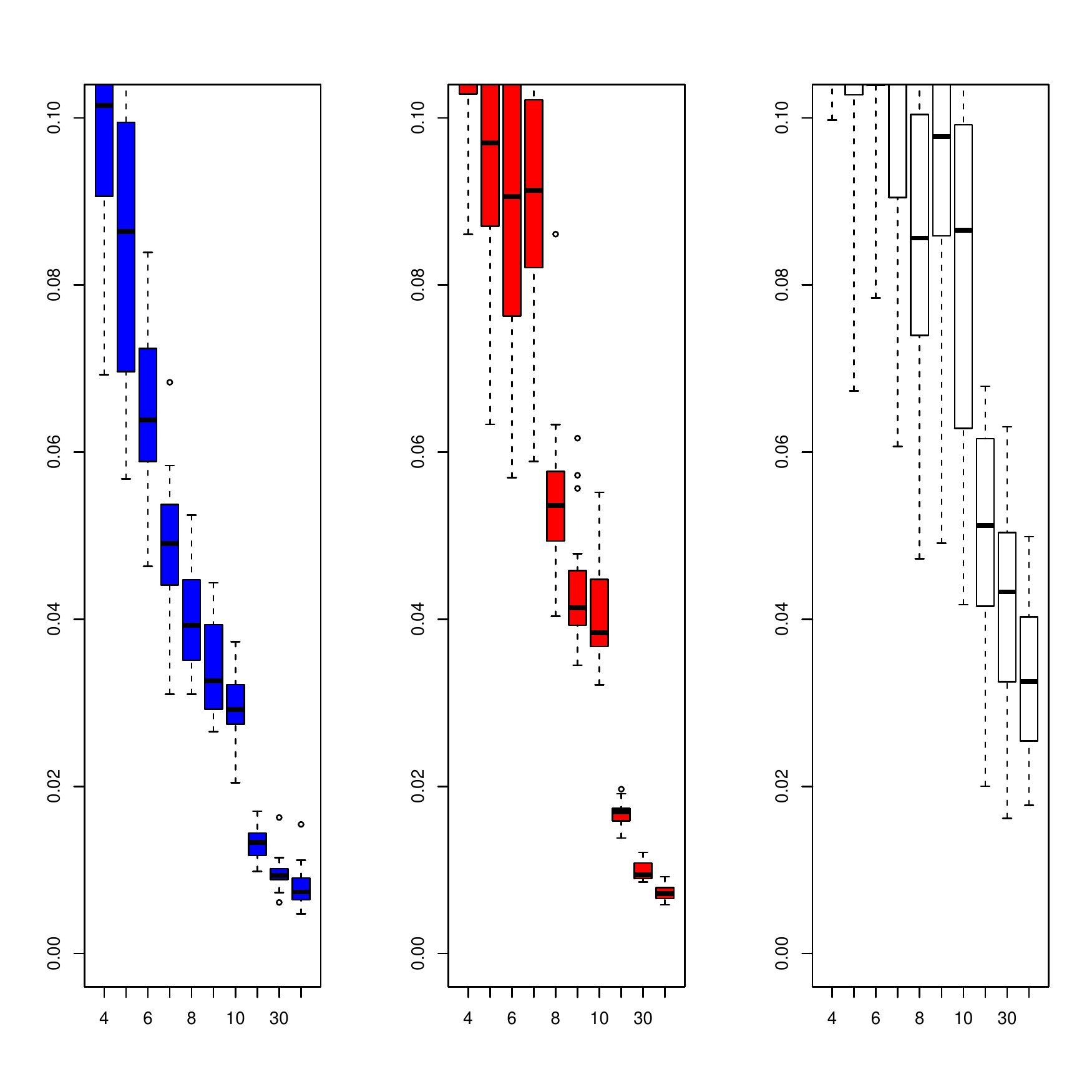} 
\includegraphics[width=0.49\textwidth]{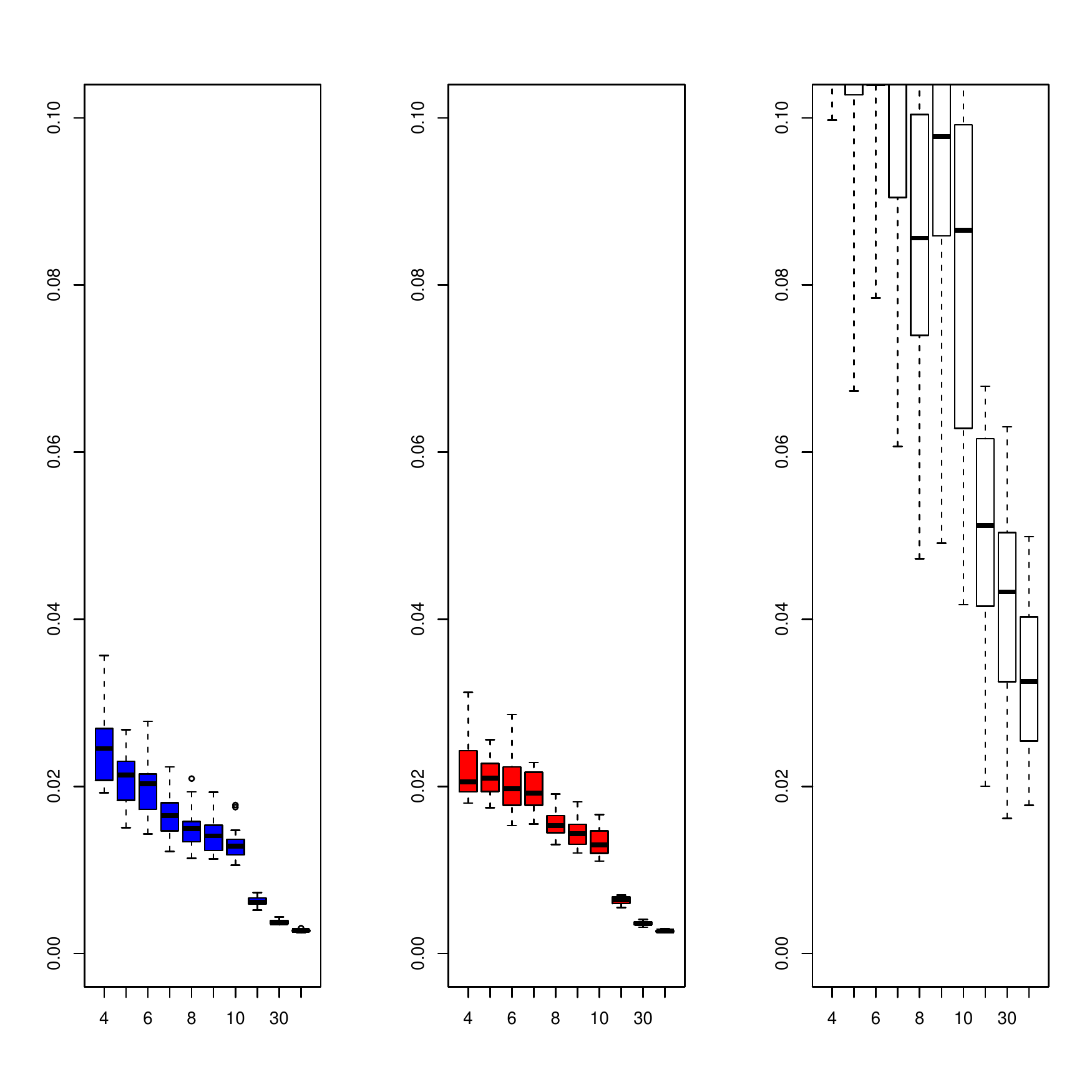} 
\vspace{-0.5cm}
\end{center}
\caption{\small{\it Top-Left:} Ten samples of FD obtained from a random functional of a Brownian bridge. 
 {\it Top-Middle-Left:} Fourier approximations based on 30 Fourier basis functions. 
 {\it Top-Middle-Right:} Piecewise constant orthonormal basis approximations based on 30  basis functions.
  {\it Top-Right:} Splinet basis approximations on 30  basis functions, the splinet is given in Figure~\ref{fig:OB}~{\it (left)}.\\
  {\it Bottom:} The boxplots of AMSE's obtained from 20 Monte Carlo simulations for the model as a function of the base size, which ranges from 4  to 40. 
 10 FD were simulated and the orthonormal basis decomposition was run with an increasing number of basis elements.  
 In the first and the fourth plot (blue), a new basis is selected anew for each MC sample, while in the second and the fifth (red) a basis is selected only for the original sample and then used for every new MC sample.  
The first two pictures correspond to the piecewise constant data-driven basis, the third and the fourth to the splinet basis, the remaining two are plots corresponding to the Fourier basis applied to the MC data.}
\label{betasample}
\vspace{-0.4cm}
\end{figure}

We illustrate, through simulations, how using a data-driven orthonormal basis can improve efficiency in representing FD.
The setting of a Monte Carlo experiment mimics, in a simplified manner, physical systems.
We use a simplified model to focus on the features of the method, while an actual physical model is discussed in the next section. 
Here, the FD are obtained by sampling parameters of the model to which we also inject samples of the Brownian bridge $\BB_i(\cdot)$:
$$
y_i(t) = F(t;\BB_i(\cdot),\theta_i), ~i=1,\dots,n.
\vspace{-.1cm}
$$  
Ten samples of FD from such models are presented in Figure~\ref{betasample}~{\it (Top-Left)}. 
The remaining graphs in Figure~\ref{betasample}~{\it (Top)} show a 40-dimensional approximation of functional signals by the Fourier, piece-wise constant, and splinet basis, respectively. The last two bases have been obtained using the DDK selection described in the previous section.

We performed orthogonal projections to the three ON bases: Fourier, piecewise constant, and the splinet, the latter is shown in Figure~\ref{fig:OB}~{\it(left)}.  
In Figure~\ref{betasample}, we show approximations of the FD that uses $N=30$ basis functions.
A Monte Carlo study of the dependence of the average mean square error on the number basis elements is shown in Figure~\ref{betasample}~{(bottom)}.
For the simulated FD, approximations were obtained with the number of basis elements increasing from 4 to 50. Their average mean square error (AMSE) over all 10 elements of the data evaluated. 
This procedure has been repeated independently 20 times resulting in 20 AMSE's for each size of the base used. 
Boxplots of these data for each model, each the base size, and each of the basis is presented in Figure~\ref{betasample}~{(bottom)}.

\section{Application -- a functional model for the extreme events}
\label{QVM}
Formally FD can be viewed as samples from a stochastic process. 
However, traditionally one refers to a stochastic process in the context of an observed individual trajectory, i.e. a single functional data sample, which is not sufficient to utilize the FDA methods. 
One can create more FD samples from a single trajectory by considering its fragments through a certain sampling process.  
Such extraction can be made through sampling at random events and extreme events are typically considered most `interesting'. 
For example, one can consider a crossing of a high level as the sampling instant and extract a sample around such an event. 
The crossing level does not need to be high to represent an extreme event -- the local maxima and minima are sampled by the derivative of the process crossing the zero level. 

For the crossing events of stationary Gaussian processes, the Slepian model has been developed to represent the stochastically mechanism that produces such data. 
For a crossing level $u$,  it has the form
\begin{equation}
\label{GaussSl}
X_u(t)=u\,r(t) - R\dot{r}(t) +\Delta(t),
\end{equation}
where  $r(t)$ is the covariance function of $X$ normalized so that $E[X^2(t)]=E[X'^2(t)]=1$,  $R$ is a standard Rayleigh variable independent from a non-stationary Gaussian process $\Delta$ having covariance
\[
r(t,s)=r(t-s)-r(t)r(s)-r'(t)r'(s).
\]
See  \cite{LeadbetterLR} and \cite{LindgrenR}. 

In \cite{Podgorski:2015aa}, a novel approach to the generalized Slepian models was proposed through the Slepian-style representation of the random noise that drives the model. 
Namely, for the Gaussian moving average 
\begin{equation*}
\label{GaussMA1}
X(t)=\int_{-\infty}^\infty g(s-t)\,dB(s),
\end{equation*}
where $dB$ is the Gaussian noise, its Slepian model can be written as 
\begin{equation*}
\label{GaussMA0}
X_u(t)=\int_{-\infty}^\infty g(s-t)\,dB_u(s),
\end{equation*}
where $dB_u$ is properly defined Slepian noise, see \cite{Podgorski:2015aa} for details.
More importantly, for any vector of stochastic processes $\mathbf Y(t)=(Y_1(t),\dots,Y_n(t))$, $t\in \mathbb R$ such that they arise as a result of some functionals acting on $B$:
\begin{align*}
Y_i(t)=H_i(t,B),~i=1,\dots,n,
\end{align*}
where for a given trajectory $B=b$, $H_i(t,b)$ can be random but independent of $B$ (and thus independent of $X'(0)$ and $X(0)$).
Then the joint Slepian model $\mathbf Y_u(t)$ for $\mathbf Y(t)$ at the instants when the moving average process $X(t)$ up-crosses level $u$ is obtained by considering
\begin{align*}
Y_{u,i}(t)=H_i(t,B_u),~i=1,\dots,n.
\end{align*} 
For example, if we consider a Gaussian moving average of the form
$$
Y(t)=\int f(s-t)~dB(s).
$$
or a non-Gaussian moving average obtained by using  the same Brownian motion $B$ subordinated by a L\'evy time change $\Gamma$,
$$
Y(t)=\int f(s-t)~dB\circ \Gamma(s)
$$
the Slepian model of $Y$ at the $u$-level crossing of $X$ is simply obtained by replacing $B$ by $B_u$. 

The approach extends to the crossing by a non-Gaussian moving average process. 
Let us consider the Laplace motion that is obtained through the subordination of $B$ by a gamma motion $\Gamma$. 
For a kernel $f$ and the L\'evy process $\Gamma$ such that $\Gamma(1)$ has the gamma distribution with shape $\tau$ and scale $1/\tau$ (for negative $t$, the process $-\Gamma(t)$ is an independent copy of $\Gamma(t), t\ge 0 $), we define the Laplace moving average (LMA)
$$
X(t)=\int g(s-t)~dL(s),
$$
where $L(t)=B(\Gamma(t))$.
Although evaluation of the crossing distributions for non-Gaussian processes is by far much more complex than those for the Gaussian process,  the principle of using the Slepian noise process can be validated and a complex high-dimensional Gibbs sampler of such a noise for the case of the Laplace moving average has been developed in  \cite{Podgorski:2015aa}. Given that the process $X$ crosses the level, one can obtain the Slepian noise $dL_u$ that has rather complex non-Gaussian distribution and simulation of it requires an advanced algorithm that has to be performed separately for each level $u$. The advantage of such an approach is that once the model for the noise is developed an arbitrary number of stochastic processes that have originally depended on the regular noise can be simply studied at the crossing events by having the original noise replaced by the Slepian noise. 
The difficulty in studying of non-Gaussian models has been reduced to a simulation of the Slepian noise. 
The goal in this section is to apply our method and find the simpler functional representation of the Slepian model. The benefit of such a simple model, is that in most applied problems, it is often sufficient to find an approximate but simple model. The simplicity of such a model can benefit numerical investigations of highly dimensional engineering problems. 
Here we show how our basis selection process can be used for the purpose, while a detailed analysis of the accuracy and efficiencies are left for some future research. 
  
\subsection{A vehicle response to an extreme event}
\begin{figure}[t!]
\begin{center}
\raisebox{-1.3cm}{
 \mbox{\includegraphics[width=0.4\textwidth]{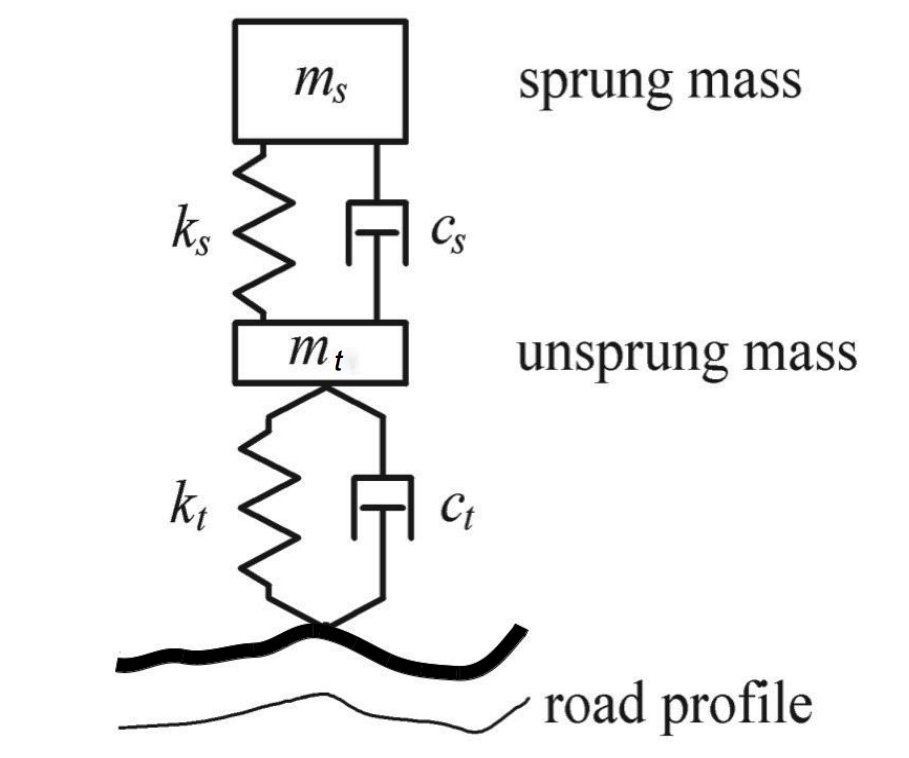}}}\hspace{1cm}
 \begin{tabularx}{4cm}{llll}
 \toprule[0.35mm]
   Parameter & Mean & Unit \\
   \midrule
   $m_s$ & 3400 & kg \\
   $k_s$ & 270 000 & N/m \\
   $c_s$ & 6000 & Ns/m \\
   $m_t$ & 350 & kg \\
   $k_t$ & 950000 & N/m \\
   $c_t$ & 300 & Ns/m\\
   \bottomrule[0.35mm]
 \end{tabularx}
 \end{center}
\caption{ Quarter vehicle model and its parameters.
}
\label{Fig:01C}\vspace{-0.5cm}
\end{figure}

For an illustration of the method we have chosen the model of a damped harmonic oscillator \eqref{eq:damped} often utilized in studies of the durability of vehicle components in the vehicle response to the road profile.
The road profile roughness is often quantified using the response of a quarter-vehicle model, see Figure~\ref{Fig:01C}, traveling at a constant velocity through road profiles. 
Such a simplification of a physical vehicle cannot be expected to predict loads exactly, but it will highlight the most important road characteristics as far as durability is concerned.
The force acting on the sprung mass $m_s$ (total mass of the vehicle) that is randomly distributed around some specific mean value is chosen as the response $Y(x)$ from the tire which then is used to compute suitable indexes to classify the severity of road roughness. 

In a linear simplification of the problem, the entire stochastic system is defined by the road elevation $R(x)$ at the location $x$ (in $[m]$) that, under constant speed $v$ of the vehicle, linearly drives two damped harmonic oscillators, one representing the tire and the other the wheel suspension system
 \begin{align*}
m_t{U}''+c_t{U}'+k_tU=dR, & \qquad
m_s{X}''+c_s{X}'+k_sX=dU.
\vspace{-.2cm}
\end{align*} 
Then $X(x)$ is the position of the mass $m_s$ (the center of the vehicle), $U(t)$ is the position of the mass $m_t$ (the `top' of the tire), and the response is $Y(x)=m_sU''(x)$. 
The vehicle parameters $\theta=(m_t,c_t,k_t,m_s,c_s,k_s)$ in the model can be set to mimic heavy vehicle dynamics, see Fig.~\ref{Fig:01C}.
They have the following physical interpretation: properties of the tire are described by $k_t$, $c_t$, which relate to stiffness and damping of the tire, while properties of the suspension are given by corresponding $k_s$, $c_s$.  
Modeling of true loads acting on components is difficult since tires filter nonlinearly the road profile and the filter parameters depend on very uncertain factors, e.g. tire's pressure, wear, etc. 
One way to account for the latter and simplify the former is to assume that some of the parameters are random and represent properties of the tire in a concrete vehicle on a given trip. 
Moreover, if a fleet of vehicles is considered the parameters can be considered random from a certain population distribution. The functional models involving this additional randomness can be also treated as discussed below although the topic will not be investigated here and from now we consider a non-random fixed $\theta$.

\begin{figure}[t]
 	\includegraphics[width=\linewidth]{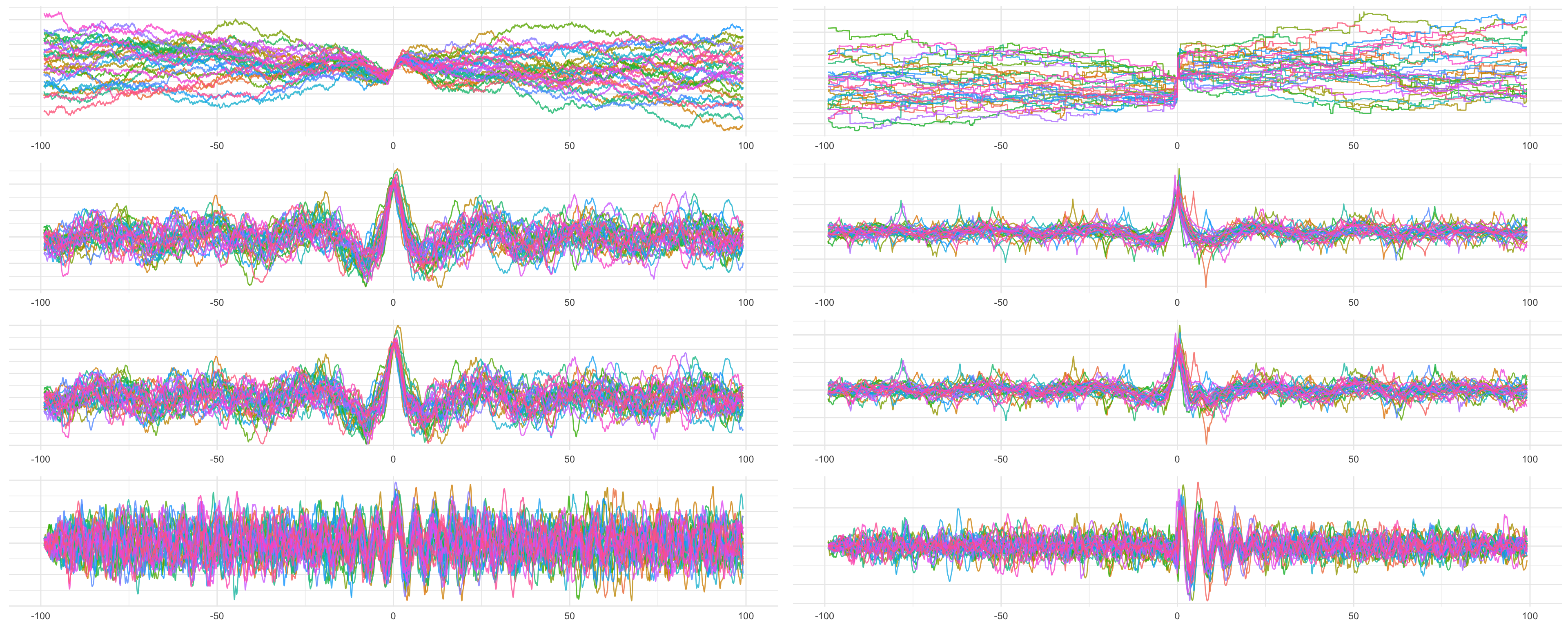}
	\includegraphics[width=\linewidth]{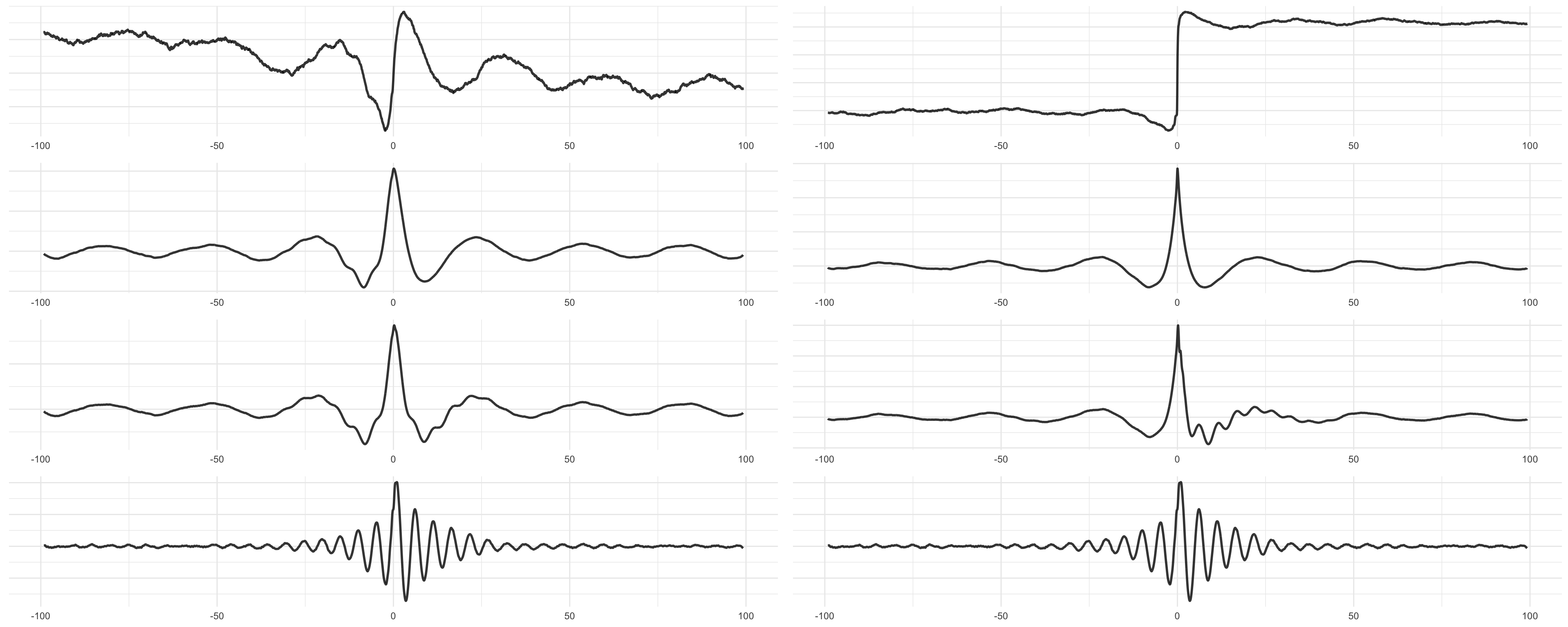}
	\includegraphics[width=\linewidth]{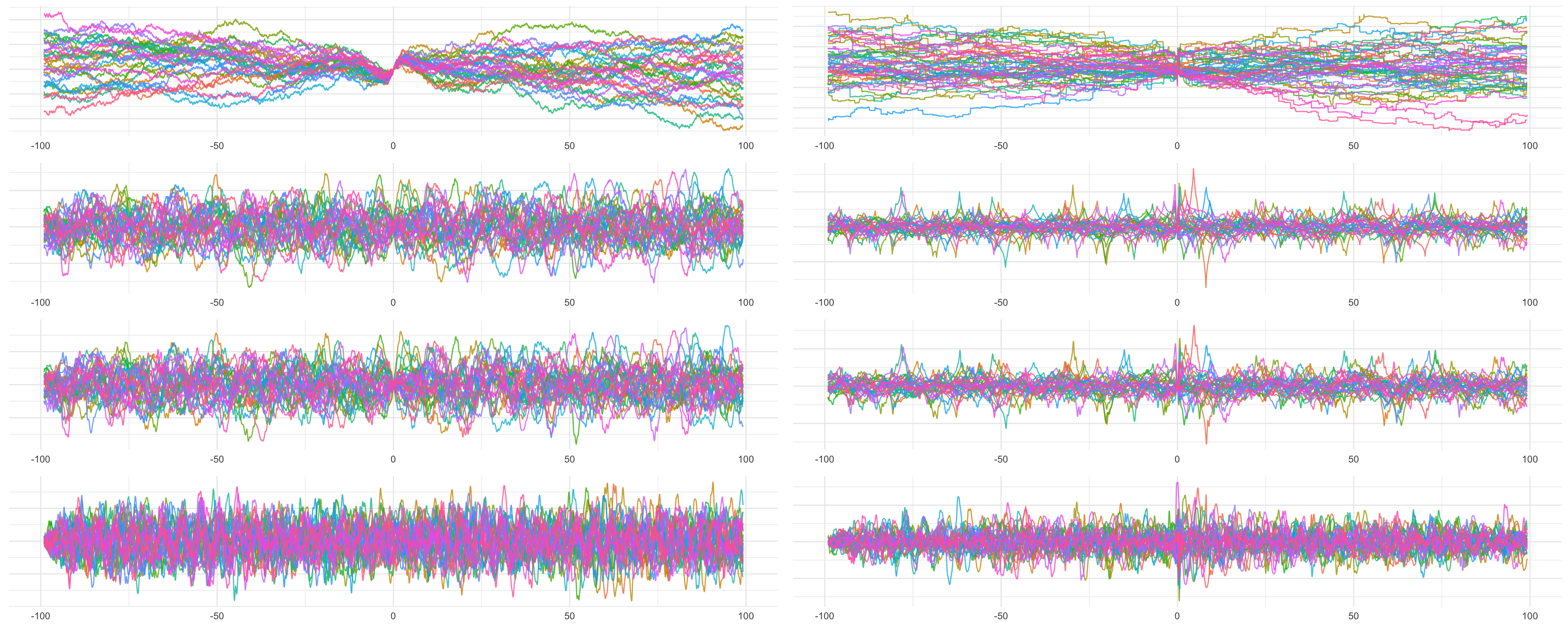}
  \caption{Slepian models for the quarter vehicle model. {\it (Top)}: 100 samples from $X_u$ and $R_u$ and $Y_u$, the three first graphs represent the Gaussian case $u=4.5$, the remaining three graphs represent the Laplace case $u=7$: {\it (Middle)}: the averaged values from the Slepian model based on 1000 simulated trajectories; {\it (Botton)} The centered samples to be modeled through a functional model.     }
\label{fig:Slepian}
\end{figure}

It is of interest to study and model both the response and the road profile at locations when $X$ reaches some extreme level. 
In what follows, we shall obtain the functional models for the Slepian noise when $X$ up-crosses $u$ that can be used to obtain Slepian models for $X_u$ and $R_u$ and $Y_u$.
Both the Gaussian and Laplace moving average models are considered with the same mean and power spectral densities and hence variances. 
The Laplace model of a road surface has a distributional parameter that can account for a large kurtosis in the data, which is also observed in empirical data. 
A typical value of the kurtosis for the type of considered roads in Sweden is 5. 
We are interested in the properties of extremal episodes when the wheel
position reaches an extreme level.
More specifically, we consider the level that is crossed about once per 600 km for the modeled stationary road.
This yields value $u=4.5$ for the Gaussian model and $u=7$ for the Laplace model. 
In Figure~\ref{fig:Slepian}~{\it (left)}, we see the output from the simulations of the Slepian models both in the Gaussian (top three graphs) and the Laplace (bottom three graphs) models.

%
%
%
In further simplification, the condition of the trip can be modeled by a Brownian bridge $\BB$ filtered by a certain kernel $r$.
Here the Brownian bridge model reflects small roughness of the road at the beginning and at the end of a trip and an increase of it when the vehicle enters tougher terrain in the middle of the trip.
The smoothing kernel $r$ represents road specific properties so that $R(t)=r*dBB(t)$ is the road surface elevation at location $t$.
In the literature, many models for the power spectral density $S_R$ of road profiles have been proposed, see \cite{Andren} for a review. 
Here, the kernel $r$ relates to the $S_R$ through the Bochner theorem $r*\tilde r(t)=2 \int \cos(\omega t) S_R(\omega)~d\omega$. 

Often one chooses the force acting on the sprung mass as the response $y(t)$ which then is used to compute suitable indexes to classify the severity of road roughness. 
In the above simplification, this response is linearly driven by the road profile, as it is also the displacement $x(t)$ of the center of the wheel from the road. 
Their transfer functions, i.e. the Fourier responses to Dirac's delta, are explicit functions of the transfer functions of the two harmonic oscillators \vspace{-0.2cm}
\[
H_t(\omega)=-m_t\,\omega^2 +i\,\omega\,c_t+k_t, \qquad H_s(\omega)=-m_s\,\omega^2 +i\,\omega\,c_s+k_s.
\vspace{-.2cm}
\]
To recap, the model is completely defined by the vehicle related parameters: the speed of the vehicle $v$, the mass of the vehicle $m_s$, the undamped angular frequencies $\omega_s$ and $\omega_t$, and the road related parameters, that describe $S_R$. All these parameters can be collectively described as $\theta$.  
Some of these parameters can be considered as random and each observed journey of a vehicle produces a response $y_i(t)$, $t\in [0,1]$, with stochastic response driven by samples of Brownian bridge $\BB_i$ and random sample $\theta_i$ of the parameters, $i=1,\dots,n$, where $n$ is the number of trips.

\begin{figure}[t!]
\begin{center}
\includegraphics[width=0.49\textwidth]{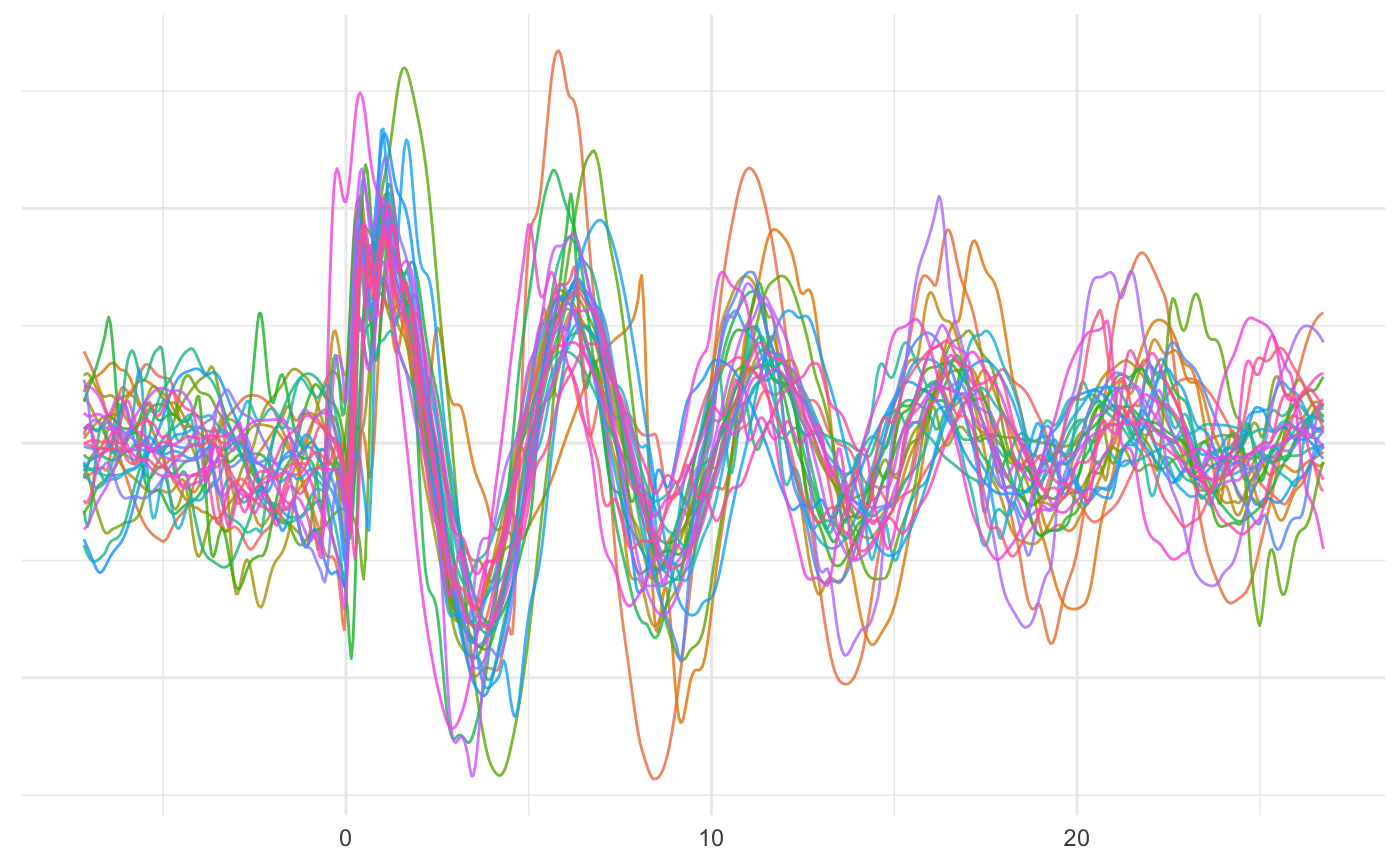}\vspace{-0.4cm}
\includegraphics[width=0.49\textwidth]{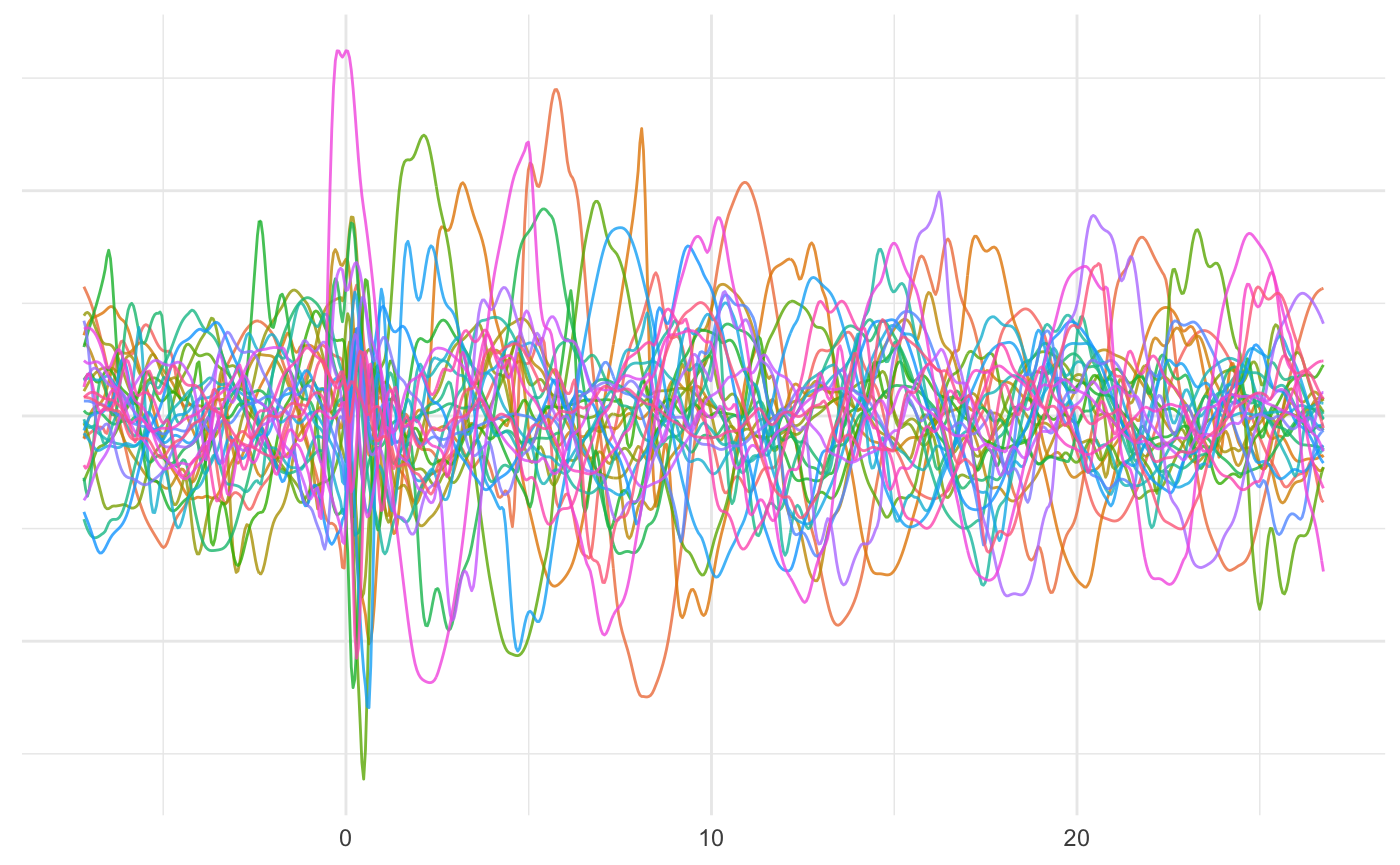}\\
\includegraphics[width=0.49\textwidth]{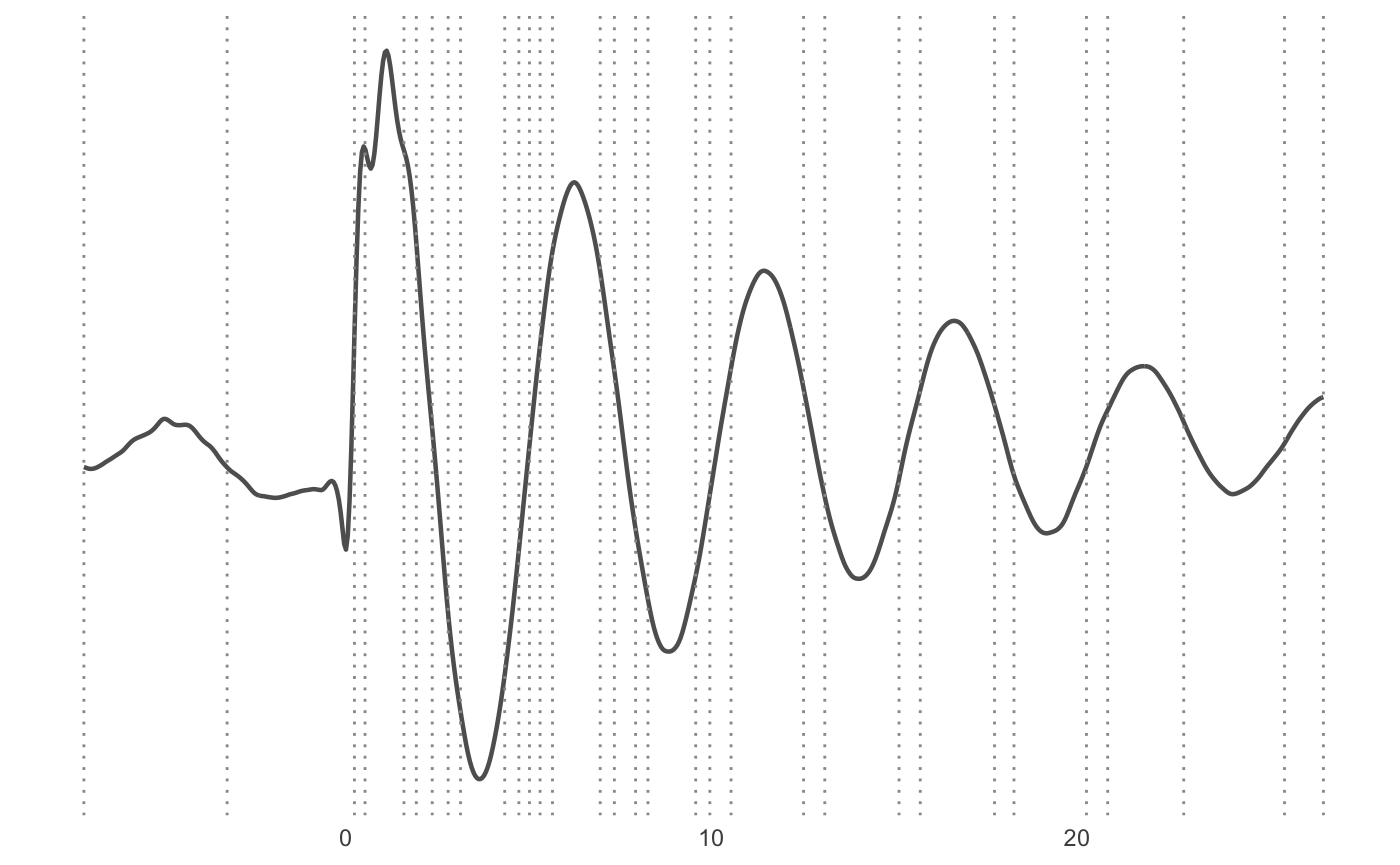}
\includegraphics[width=0.5\linewidth]{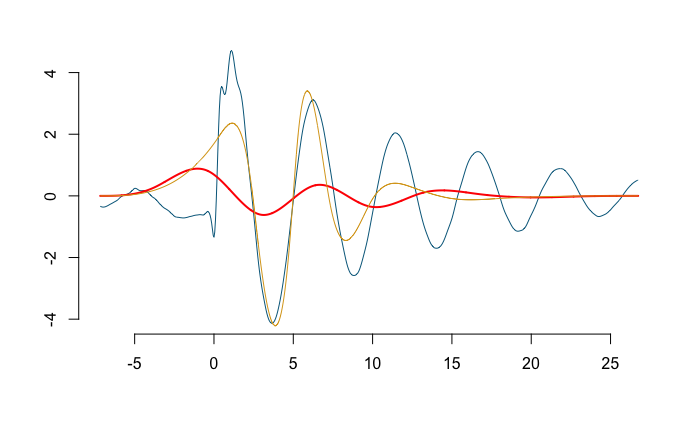}
\end{center}
\caption{\small {\it Top-Left:} 30 samples from the Truck date (Laplace case) when reaches some extreme level;  {\it Top-Right:} its centered samples.
{\it Bottom-Left:} The sample mean with 30 selected knots shown through vertical dashed lines.
{\it Bottom-Right:}
The sample mean in the time interval $[-7,26]$ (blue curve), its splinet representation build over 10 DDK selected knots (orange curve) and its splinet representation build over 10 equally spaced knots (red curve).
}
\label{fig:LtruckSub} \vspace{-0.1cm}
\end{figure}

\subsection{The DDK decomposition near an extreme event}
\label{sec:SS}
In the sequel, we will apply the knot section algorithm on the Slepian noise model for the Laplace case-truck data.  Our focus is to retrieve local feature near the origin which is located where the extreme event occurs for this reason we restrict ourselves to the interval $[-7,26]$ and 30 FD on this interval are shown in Figure~\ref{fig:LtruckSub}~{\it (Top-Left)}.

First, we show how the algorithm performs for the trend in the functional data, i.e. we use it on the averaged functions.  
Figure~\ref{fig:LtruckSub}~{(\it Bottom-Left)} shows how the knots are placed to capture functional variability in a datum on the time interval $[-7,26]$. 
Figure~\ref{fig:LtruckSub}~{(\it Bottom-Right)} presents a comparison between the performance of equally spaced knot selection and the DDK method when applied on the sample mean. 
We can clearly see the efficiency in capturing the shape of the sample mean curve when using the DDK method.



\begin{figure}[t!]
\begin{center}
\includegraphics[width=0.75\textwidth]{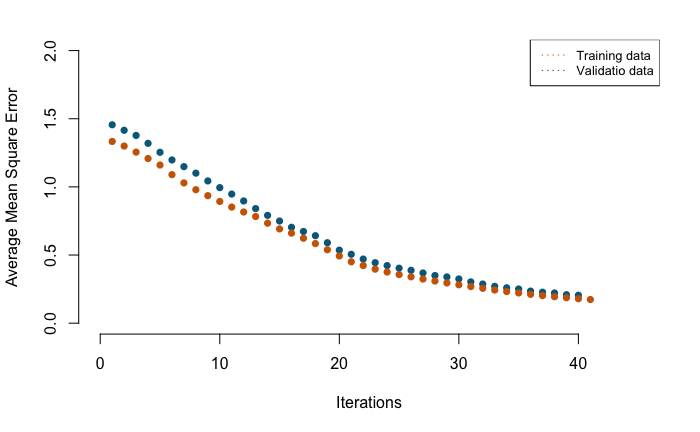}
\end{center}
  \caption{Reduction in AMSE in the Slepian noise model for the Laplace case-truck data. {\it (Orange-bottom)}  reduction in AMSE achieved after each additional knot selection during training. {\it (Blue-top)} reduction achieved after each additional knot selection on the validation data.}
\label{fig:TrainValidationErrorReduction} \vspace{-0.2cm}
\end{figure}

Next, we perform the FDA analysis on the centered data. A sample of 30 is shown in Figure~\ref{fig:LtruckSub}~{\it (Top-Right)}.
 After splitting the centered dataset randomly into training and validation dataset with a split percentage 60\%, 40\% respectively, we apply the knot selection algorithm on the train dataset and validate our choice simultaneously on the validation dataset at every iteration.  At each iteration, a new knot is selected and the average mean square error will reduce in both the train dataset and the validate dataset, as can be seen in Figure~\ref{fig:TrainValidationErrorReduction}. To determine the optimal number of knots, we select the number of knots at the “elbow”, i.e. the point after which the distortion/inertia start decreasing in a linear fashion,  in the reduction of the average mean square error in the validation data set. Thus for the given data, we decide for $25$ knots.

\begin{figure}[t!]
\begin{center}
\includegraphics[width=0.75\linewidth]{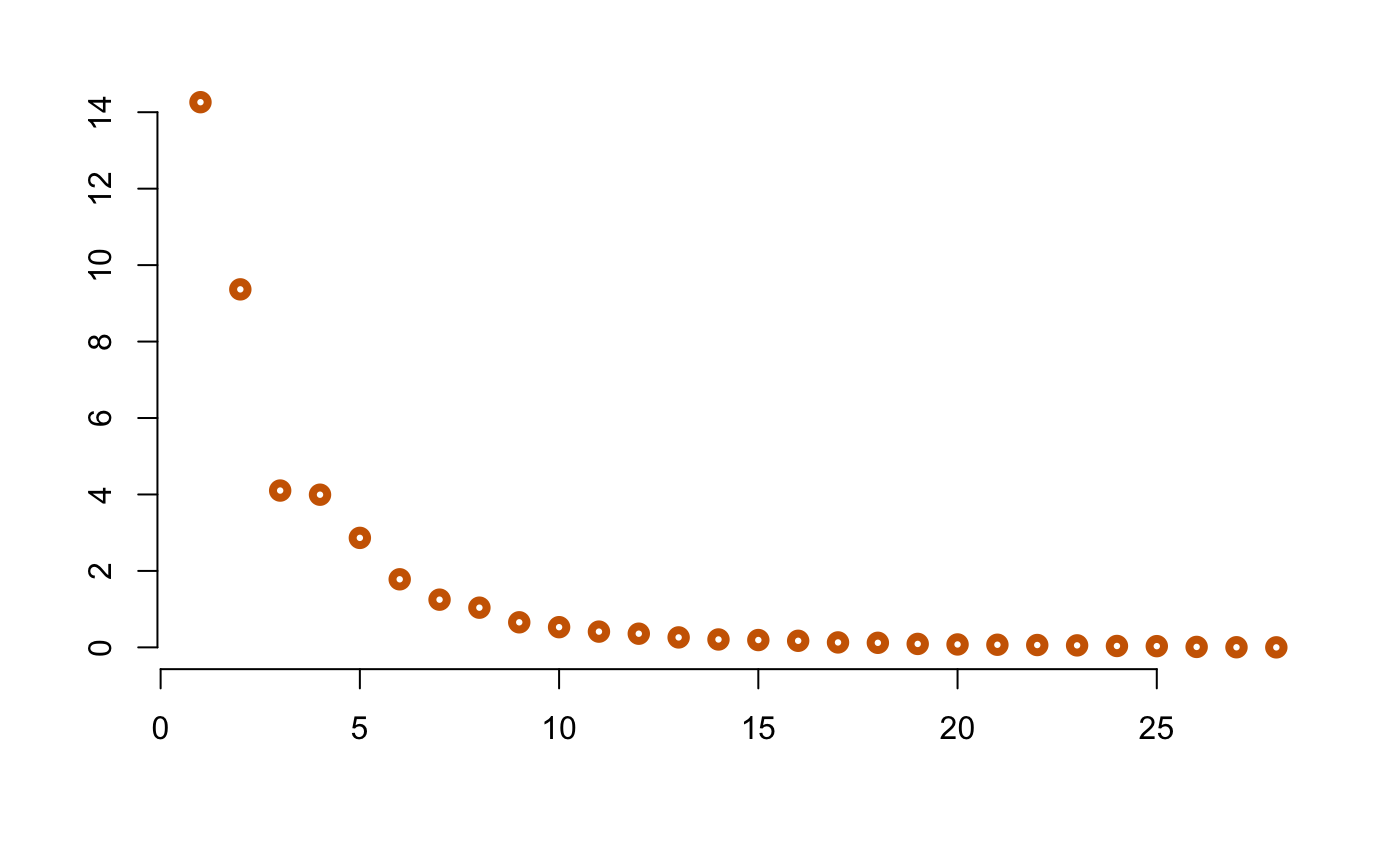}\\
\includegraphics[width=0.5\linewidth]{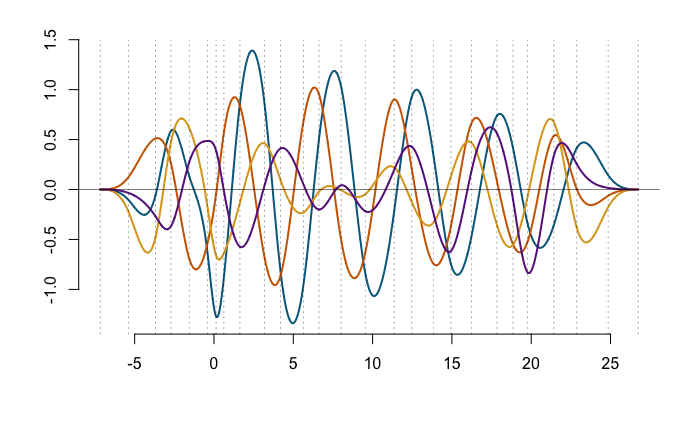}
\includegraphics[width=0.49\linewidth,]{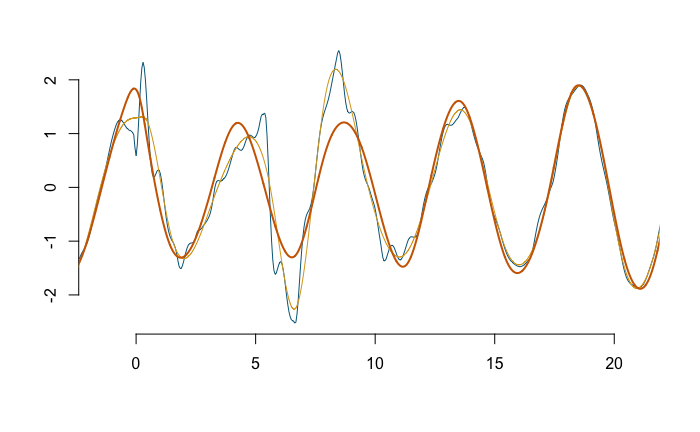}
\end{center}
  \caption{ {\it Top:} The eigenvalues ordered in a decreasing manner; {\it Bottom-Left:} the first four eigenfunctions  scaled by the square roots of their corresponding eigenvalues;
  {\it Bottom-Right:} The original data around the extreme level (Blue curve), the projected data around the extreme level into splinets build over the knots chosen with DDK (yellow curve), and the data decomposed using the first four eigenfunctions (orange curve).}
\label{fig:eigen} \vspace{-0.51cm}
\end{figure}

After selecting the knots, we use the function  {\tt project()}, from the {\tt Splinets}  R-package, to project data into splinets of the third order with the selected knots. In the next step, we perform the spectral decomposition of data by estimating the eigenvalues $\lambda_i$’s and the corresponding eigenfunctions $f_i(t)$.  We present the estimated eigenvalues in
the decreasing order and the first four eigenfunctions scaled by the square roots of their corresponding eigenvalues in Figure~\ref{fig:eigen}. The vertical dashed lines refer to selected knots.

Figure~\ref{fig:eigen}~{\it (Bottom-Right)} shows, for a  sample in the data, the difference between the original centered truck data, the projected into splinets over the knots selected using the data-driven developed approach, and the functional data spectrally decomposed and reconstructed using the first four eigenfunctions. It is apparent that the projection using the DDK method gives a decent smooth 4D fit of the original data and efficiently avoids overfitting the original curve.

At the last step of this study we compare between  $\hat{f}_1$, the projected data into splinets build over the DDK and  $\hat{f}_2$, the projected data into splinets build over the same number of equally spaced knots. The average mean square errors between the original data and the projected data are

$$
 AMSE(f- \hat{f}_1) = 0.2095587, \qquad \qquad AMSE(f- \hat{f}_2) =  0.2542373.
$$
Despite that due to the nature of this data the locality occurs only near the extreme event, we observe significant improvement of the quality (by 20\%) of the averaged mean square error.

In the conclusion of this study, we report that a complex non-Gaussian model of a vehicle response $Y_u$ near an extreme event that requires a very complex sampling techniques to obtain an accurate process can be adequately approximated by a simple low-dimensional model
$$
\tilde Y_u(t)= \mu(t) + \sqrt{\lambda_1} Z_1 f_1(t)+\sqrt{\lambda_2} Z_2 f_2(t)+\sqrt{\lambda_3} Z_3 f_3(t)+\sqrt{\lambda_4} Z_4 f_4(t),
$$
in which the distribution of the vector standardized uncorrelated random variables $(Z_1,Z_2,Z_3,Z_4)$ can be fitted by taking the inner products of the data with the eigenfunctions and standardized obtained data. 
Our inspection of this empirical distribution showed that it does not deviate in a drastic manner from the standard vector normal distribution. Thus in further simplification of the model, one can obtain samples from it using samples of standard normal variables. 
It is clear that this simplistic model is not providing insight to all complexities that may be present in the physical system, however, it could be used as a proxy to obtain fast data to roughly assess the behavior of a vehicle at an extreme event.

\vspace{-.4cm}
\section*{Conclusions}
\vspace{-.5cm}
\begin{svgraybox}
The proposed method of the data-driven orthonormal basis decomposition has been tested both through numerical simulations and on vehicle mechanics data. 
The Monte Carlo simulations show clear advantages over the Fourier based method, in particular, when smoothed splines are used. 
The accuracy is not only exhibited in smaller errors but also in the reduced variability of the error. 
The improvement is greater, as expected, for the data that shows some local detail. 
The obtained results suggest that the method may have the potential to improve the functional analysis of the data coming from physical systems with random excitation and involving random parameters as shown through the data of vehicle response to a transient in a road surface. 
The true benefits of the proposed methodology exhibit for the sparse data which was demonstrated through Monte Carlo simulations. 
In the future, the method will be applied to empirical massive sparse data such as LC-MS/MS spectra, where the main feature of the method, the efficiency in handling sparse data, is expected to bring a significant improvement to statistical analysis of the so-called  {\it omics data}. 
\end{svgraybox}

\printbibliography

@article{Liu2019SplinetsE,
	Author = {X. Liu and H. Nassar and K. Podg{\'o}rski},
	Journal = {ArXiv},
	Title = {Splinets - efficient orthonormalization of the B-splines},
	Volume = {abs/1910.07341},
	Year = {2019}}

@article{LindgrenR,
	Author = {Lindgren, G. and Rychlik, I.},
	Journal = {International Statistical Review},
	Pages = {195-225},
	Title = {Slepian models and regression approximations in crossing and extreme value theory},
	Volume = {59},
	Year = {1991}}

@book{LeadbetterLR,
	Author = {Leadbetter, M.R. and Lindgren, G. and Rootzen, H.},
	Publisher = {Springer-Verlag},
	Title = {Extremes and related properties of random sequences and processes},
	Year = {1983}}

@article{GuoHJZ,
	Author = {Guo, J. and Hu, J. and Jing, B.-Y. and Zhang, Z.},
	Journal = {Journal of the American Statistical Association},
	Number = {513},
	Pages = {288-297},
	Title = {Spline-Lasso in High-Dimensional Linear Regression},
	Volume = {111},
	Year = {2016}}

@article{Andren,
	Author = {Andr{\'e}n, P.},
	Journal = {Int. J. Vehicle Design},
	Pages = {2-14},
	Title = {Power spectral density approximations of longitudinal road profiles.},
	Volume = {40},
	Year = {2006}}

@article{Podgorski:2015aa,
  title={Slepian noise approach for gaussian and Laplace moving average processes},
  author={Podg{\'o}rski, K. and Rychlik, I. and Wallin, J.},
  journal={Extremes},
  volume={18},
  number={4},
  pages={665--695},
  year={2015},
  publisher={Springer}
}

@misc{ref:Podgorski,
	author = {Podg\'orski, K.},
	howpublished = {<arXiv:2102.00733>.},
	title = {Splinets -- splines through the Taylor expansion, their support sets and orthogonal bases.},
	year = {2021}}

@book{hsing,
	Author = {Hsing, T. and Eubank, R.},
	Publisher = {John Wiley \& Sons},
	Title = {Theoretical foundations of functional data analysis, with an introduction to linear operators},
	Year = {2015}}

@article{ramsay2004functional,
	Author = {Ramsay, J.O.},
	Journal = {Encyclopedia of Statistical Sciences},
	Publisher = {Wiley Online Library},
	Title = {Functional data analysis},
	Volume = {4},
	Year = {2004}}

@book{ferraty2006nonparametric,
	Author = {Ferraty, Fr{\'e}d{\'e}ric and Vieu, Philippe},
	Publisher = {Springer Science \& Business Media},
	Title = {Nonparametric functional data analysis: theory and practice},
	Year = {2006}}

@book{Deboor,
	Author = {De Boor, C.},
	Edition = {revised},
	Publisher = {Springer-Verlag New York},
	Series = {Applied Mathematical Sciences},
	Title = {A practical guide to splines},
	Volume = {27},
	Year = {2001}}

@book{HastieTF9,
	Author = {Hastie, T. and Tibshirani, R. and Friedman, J.},
	Edition = 2,
	Publisher = {Springer},
	Title = {The elements of statistical learning: data mining, inference and prediction},
	Url = {http://www-stat.stanford.edu/~tibs/ElemStatLearn/},
	Year = 2009}

@article{Karhunen,
	Author = {Karhunen, K.K.},
	Journal = {Ann. Acad. Sci. Fennicae. Ser. A.},
	Pages = {1-79},
	Title = {\"Uber lineare Methoden in der Wahrscheinlichkeitsrechnung},
	Volume = {37},
	Year = {1947}}

@inproceedings{EDOB,
  title={Empirically Driven Orthonormal Bases
for Functional Data Analysis},
  author={ H. Nassar and K. Podg{\'o}rski},
  booktitle={Numerical Mathematics and Advanced Applications ENUMATH 2019, Lecture Notes in Computational Science and Engineering 139},
  pages={1--12},
  year={2021},
  organization={Springer}
}


\end{document}